\pdfoutput=1

\documentclass[11pt]{article}

\usepackage[preprint]{acl}

\usepackage{times}
\usepackage{latexsym}
\usepackage{amsmath, amssymb} 
\usepackage{graphicx}         
\usepackage{booktabs}         
\usepackage{xcolor}           
\usepackage{algorithm}        
\usepackage{algorithmic}      
\usepackage{listings}         
\usepackage{hyperref}         
\usepackage{multirow}
\usepackage{makecell}
\usepackage{enumitem}
\usepackage{lineno}
\usepackage{pgfplots}
\usepackage{pgfplotstable}
\usepgfplotslibrary{groupplots}
\pgfplotsset{compat=1.18}
\usepackage{fancyvrb}
\usepackage{url}

\usepackage[T1]{fontenc}

\usepackage[utf8]{inputenc}

\usepackage{microtype}

\usepackage{inconsolata}

\usepackage{graphicx}

\title{Selecting Demonstrations for Many-Shot In-Context Learning\\via Gradient Matching}

\author{
    Jianfei Zhang\textsuperscript{1},
    Bei Li\textsuperscript{2},
    Jun Bai\textsuperscript{3},
    Rumei Li\textsuperscript{1,2},
    Yanmeng Wang\textsuperscript{4},
    Chenghua Lin\textsuperscript{5},
    Wenge Rong\textsuperscript{1}\\
    \textsuperscript{1}School of Computer Science and Engineering, Beihang University, China \\
    \textsuperscript{2}Meituan, Inc., China ~  \textsuperscript{3}Beijing Institute for General Artificial Intelligence, China \\
    \textsuperscript{4}Ping An Technology, China ~  
    \textsuperscript{5}Department of Computer Science, University of Manchester, UK \\
    \texttt{\{zhangjf, lirumei3232, w.rong\}@buaa.edu.cn}, ~ \texttt{libei17@meituan.com} \\
    \texttt{baijun@bigai.ai}, ~
    \texttt{wangyanmeng219@pingan.com.cn}, ~
    \texttt{chenghua.lin@manchester.ac.uk}
}

\begin{document}
\maketitle
\begin{abstract}
In-Context Learning (ICL) empowers Large Language Models (LLMs) for rapid task adaptation without Fine-Tuning (FT), but its reliance on demonstration selection remains a critical challenge. While many-shot ICL shows promising performance through scaled demonstrations, the selection method for many-shot demonstrations remains limited to random selection in existing work. Since the conventional instance-level retrieval is not suitable for many-shot scenarios, we hypothesize that the data requirements for in-context learning and fine-tuning are analogous. To this end, we introduce a novel gradient matching approach that selects demonstrations by aligning fine-tuning gradients between the entire training set of the target task and the selected examples, so as to approach the learning effect on the entire training set within the selected examples. Through gradient matching on relatively small models, e.g., Qwen2.5-3B or Llama3-8B, our method consistently outperforms random selection on larger LLMs from 4-shot to 128-shot scenarios across 9 diverse datasets. For instance, it surpasses random selection by $4\%$ on Qwen2.5-72B and Llama3-70B, and by around $2\%$ on 5 closed-source LLMs. This work unlocks more reliable and effective many-shot ICL, paving the way for its broader application.
\end{abstract}

\section{Introduction}
In-Context Learning (ICL) enables pre-trained Large Language Models (LLMs) to perform tasks by learning from input-output examples (or ``\textit{demonstrations}'') provided during inference \cite{DBLP_gpt3}. This allows LLMs to adapt to new tasks through forward propagation, without weight updates via back-propagation, offering a flexible alternative to traditional fine-tuning.

\begin{figure}[t]
  \includegraphics[width=\columnwidth]{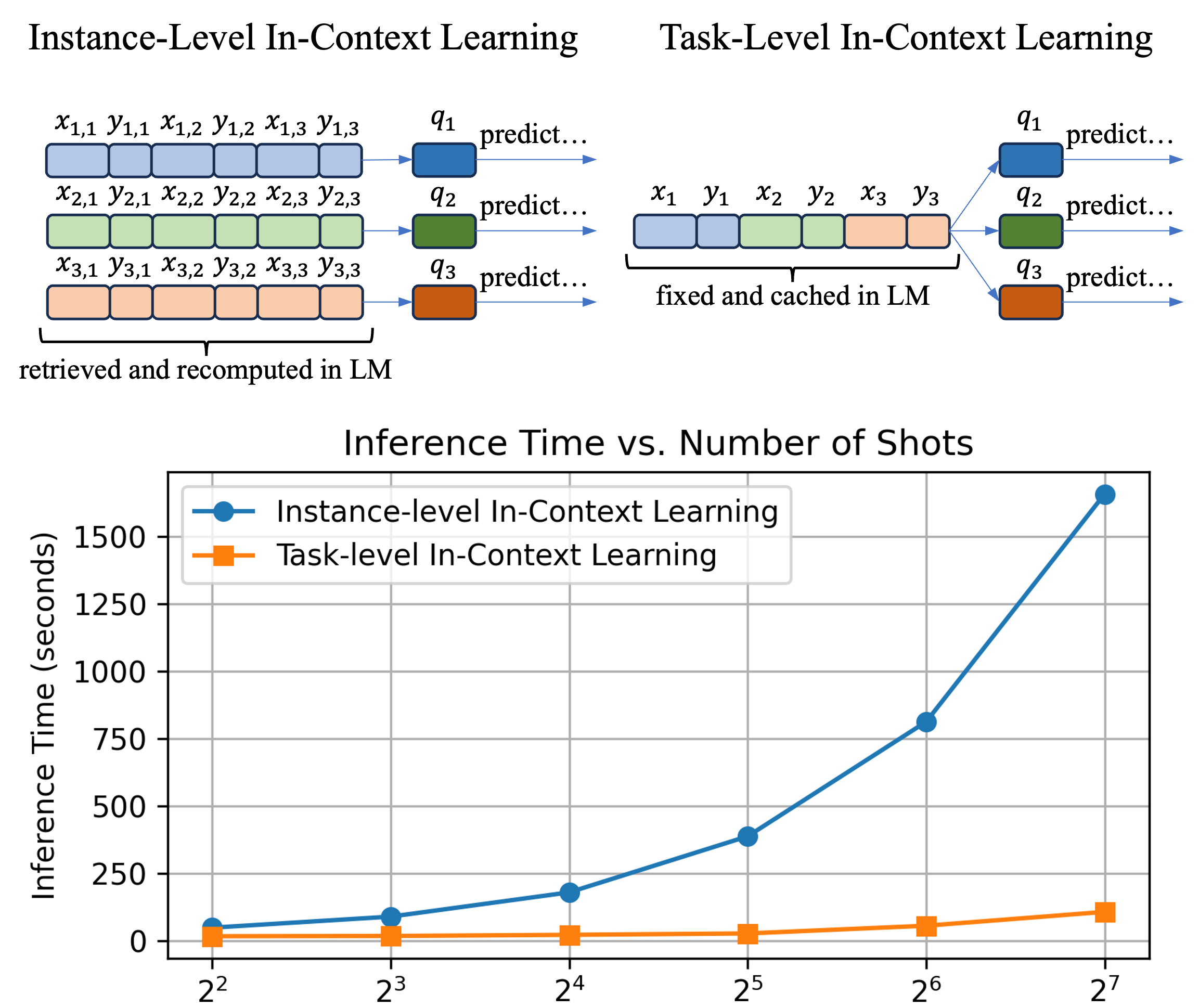}
  \caption{Instance-level retrieved demonstrations are not suitable for Many-Shot ICL, since the variable context cannot utilize prefix-caching and can lead to theoretically $O(n^2)$ inference complexity \cite{DBLP_transformer} when using $n$-shot demonstrations. Here we illustrate the runtime of Llama3-70B \cite{DBLP_llama3} on CMSQA \cite{DBLP_cmsqa} using vLLM with 4 A100 GPUs in tensor-parallel \cite{DBLP_vLLM}.}
  \label{fig:duration}
\end{figure}

Since the performance of ICL is often sensitive to the choice of demonstrations \cite{DBLP_icl_sensitive}, significant efforts have been made to improve demonstration selection. Most studies concentrate on instance-level retrieval, aiming to identify suitable demonstrations for each test query independently \cite{DBLP_icl_retrieval,DBLP_icl_retrieval2}. This mainly involves considering similarity \cite{DBLP_icl_retrieval,DBLP_icl_retrieval2} between demonstrations and the query, as well as auxiliary factors such as complexity \cite{DBLP_icl_complexity}, perplexity \cite{DBLP_icl_perplexity}, difficulty \cite{DBLP_icl_difficulty}, and diversity \cite{DBLP_icl_diversity}. Another line of work focuses on task-level selection, which seeks to find a fixed set of demonstrations that achieves the best average performance across all test queries from a target task \cite{DBLP_icl_rl,DBLP_latent_bayesian}. Current approaches primarily explore reinforcement learning on the selection policy \cite{DBLP_icl_rl} and Bayesian inference to explain the demonstration effectiveness \cite{DBLP_latent_bayesian} as potential solutions.

However, for the recently emerged Many-Shot in-context learning \cite{DBLP_manyshoticl} paradigm, demonstrations are selected simply by random in existing work~\cite{DBLP_manyshoticl,DBLP_manyshoticl_cls,DBLP_manyshoticl_verifier,DBLP_manyshoticl_multimodal}, which may lead to suboptimal performance. Nevertheless, existing methods for few-shot demonstration selection are not well-suited to many-shot scenarios. On one hand, applying instance-level demonstrations conflicts with caching and reusing the hidden states of the same long context in language models \cite{DBLP_prefix_caching}, leading to theoretically $O(n^2)$ inference complexity \cite{DBLP_transformer} for $n$-shot demonstrations, as illustrated in Fig.~\ref{fig:duration}. On the other hand, existing task-level selection methods are designed for very few-shot scenarios (e.g., 4-shot in their implementations), facing challenges from exploration complexity \cite{DBLP_icl_rl} and performance early saturation \cite{DBLP_latent_bayesian} for more shots. To further release the potential of many-shot ICL, it now requires a selection method with scalable effectiveness to many-shot scenarios.

To address such research gap, we revisit ICL demonstration selection from a ``learning'' perspective—we hypothesize that the \textit{data requirements for in-context learning and fine-tuning are analogous}. Building on this premise, we conduct latent concept learning \cite{DBLP_latent_bayesian} on a small Language Model (LM) to learn the optimal in-context task embeddings. Then we compute the latent gradient that each example provides to the learning process. Finally, we select $n$-shot demonstrations from the whole fine-tuning set with the minimized $L_2$ distance between their average latent gradients, so as to approach the learning effect on all examples within the selected $n$-shot demonstrations.

We validate our proposed method on 9 datasets from 5 distinct NLP tasks, each covering scenarios from 4-shot to 128-shot. Our method selects demonstrations through gradient matching on relatively small models, e.g., Qwen2.5-3B or Llama3-8B, and surpasses the widely-adopted random selection by an average of $4\%$ on Llama3-70B and Qwen2.5-72B. Furthermore, the selected demonstration set exhibits transferability to closed-source LLMs, consistently outperforming random selection by around $2\%$ on Qwen-turbo, GLM-4-flash, Doubao-pro-32k, GPT-4o-mini, and DeepSeek-V3. Our source code is available at \url{https://github.com/zhangjf-nlp/ManyShotICL-CLG.git}.

\section{Related Work}

\subsection{Many-Shot In-Context Learning}
In-Context Learning (ICL) refers to the capability of LLMs to learn from data during inference through forward propagation, without the need for backward propagation or weight updates~\cite{li-etal-2023-compressing}. Prior work is limited to few-shot scenarios \cite{DBLP_manyshoticl_multimodal} by the context window size, e.g., 2048 tokens in GPT-3 \cite{DBLP_gpt3}.

Recently, advancements in expanding the context window size of LLMs \cite{DBLP_longcontext1,DBLP_longcontext3}, e.g., up to 128k tokens in GPT-4o \cite{DBLP_gpt4o} and Qwen2.5 \cite{DBLP_Qwen25}, have enabled the exploration of many-shot ICL \cite{DBLP_manyshoticl}. These studies have observed that Many-Shot ICL, which includes up to hundreds or even thousands of demonstrations, can make substantial improvements compared to few-shot ICL on various tasks \cite{DBLP_manyshoticl,DBLP_manyshoticl_cls}. In this paradigm, the benefits of instance-level retrieval over using a fixed random set of demonstrations tend to diminish \cite{DBLP_manyshoticl_cls}, while a fixed demonstration set can largely reduce the inference cost through prefix-caching \cite{DBLP_manyshoticl}. Therefore, researchers tend to randomly select a demonstration set and reuse it across all queries in many-shot ICL \cite{DBLP_manyshoticl,DBLP_manyshoticl_cls,DBLP_manyshoticl_multimodal,DBLP_manyshoticl_verifier}.

While random selection is widely adopted for many-shot ICL, it is underexplored whether this strategy is truly optimal. This motivates our work to explore and develop a better demonstration selection strategy for the many-shot paradigm.

\begin{figure*}[htp]
    \centering
    \includegraphics[width=\textwidth]{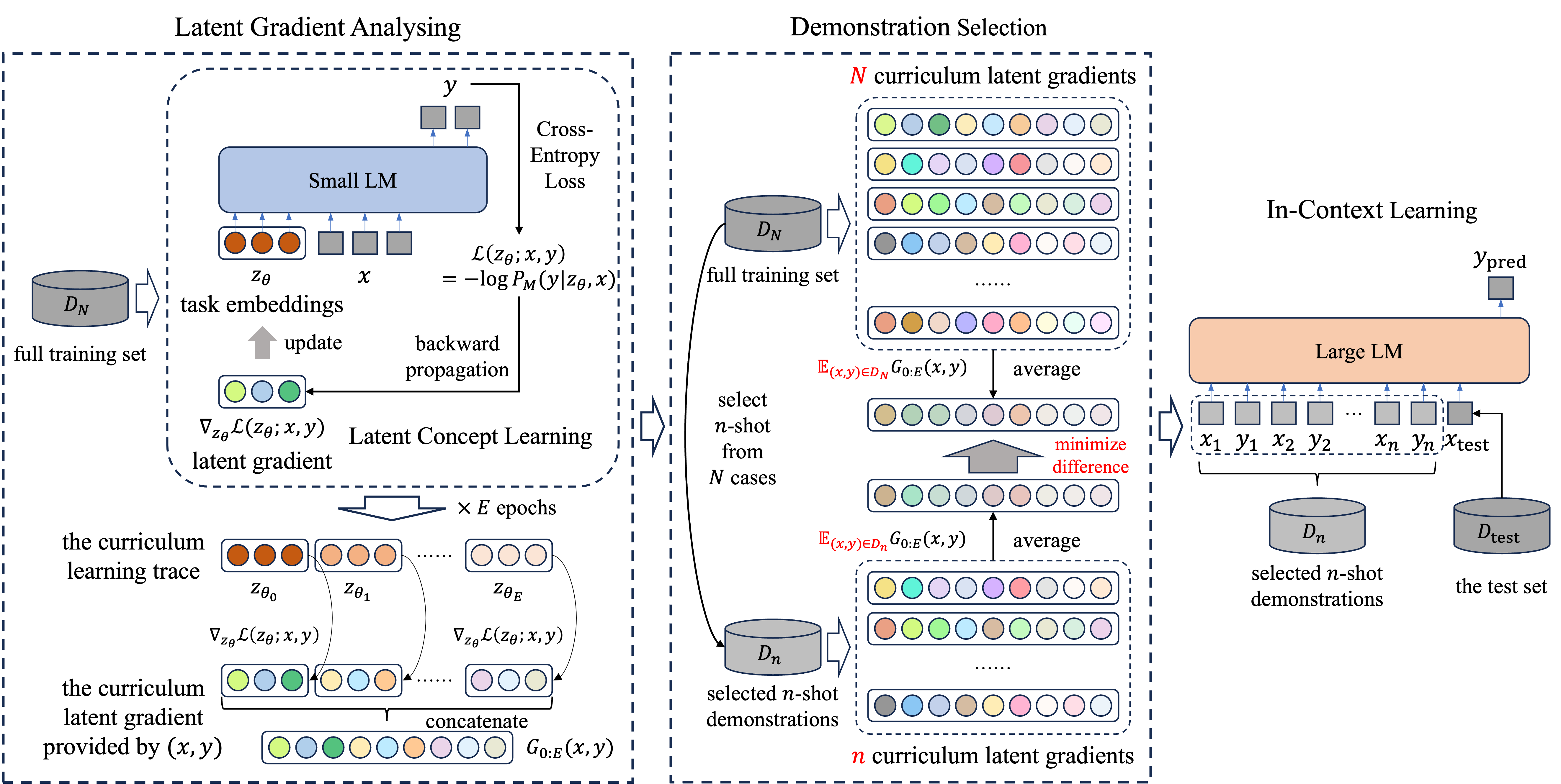}
    \caption{The overview of our proposed method for many-shot demonstration selection. We perform latent concept learning \cite{DBLP_latent_bayesian} on a small LM and compute the Curriculum Latent Gradient (CLG) provided by each example throughout the $E$-epoch training. Then we find an $n$-shot demonstration set to mimic the average gradient on the entire training set. We show that the selected $n$-shot demonstrations can effectively improve the many-shot In-Context Learning performances on both same-series larger LMs and various closed-source LLMs.}
    \label{fig:main}
\end{figure*}
\subsection{Demonstration Selection}
The performance of ICL is sensitive to the choice of demonstrations \cite{DBLP_icl_sensitive,DBLP_icl_sensitive2}, prompting efforts in demonstration selection. Existing approaches fall into two categories:

\paragraph{Instance-level retrieval} selects demonstrations for each query using semantic similarity, such as cosine similarity \cite{DBLP_icl_retrieval2} and BM25 \cite{DBLP_icl_bm25}. Some work further emphasizes additional factors including diversity \cite{DBLP_icl_diversity}, complexity \cite{DBLP_icl_complexity}, difficulty \cite{DBLP_icl_difficulty}, and perplexity \cite{DBLP_icl_perplexity}. While providing relevant and useful information, instance-level retrieval prevents the use of prefix-caching during inference and leads to theoretically $O(n^2)$ inference complexity \cite{DBLP_transformer} for $n$-shot demonstrations. In contrast, the runtime of many-shot ICL under a fixed demonstration set increases only linearly with a large number of shots \cite{DBLP_manyshoticl}. This inefficiency makes instance-level retrieval less practical as the number of demonstration shots increases.

\paragraph{Task-level selection} aims to identify a fixed set of demonstrations that achieve optimal average performance across all queries from a target task. For example, \citet{DBLP_icl_rl} use Reinforcement Learning (RL) to gradually explore the optimal demonstration set. \citet{DBLP_latent_bayesian} view LLMs as latent variable models and select demonstrations that can best infer the optimal latent task concept. However, these methods are both limited to few-shot scenarios (e.g., 4-shot in their implementations), facing challenges from complexity in exploring potential many-shot demonstration sets \cite{DBLP_icl_rl} and performance early saturation beyond 4-shot \cite{DBLP_latent_bayesian} respectively.

In summary, current demonstration selection methods are fundamentally designed for and evaluated within few-shot scenarios, facing challenges in generalizing to many-shot scenarios. To the best of our knowledge, our work presents the first attempt to enhance demonstration selection for many-shot ICL, beyond random selection in existing work.

\section{Methodology}

Previous research has identified several key attributes that contribute to the effectiveness of in-context demonstrations, such as similarity \cite{DBLP_icl_retrieval}, diversity \cite{DBLP_icl_diversity}, and coverage \cite{DBLP_icl_diversity}. While intuitively beneficial, these attributes lack a direct connection to the \textit{learning dynamics} induced by demonstrations within the language model. We posit that the ideal demonstration set for ICL should not only present static attributes of high-quality fine-tuning data but, more fundamentally, should actively facilitate and guide the language model's learning process on the target task. Drawing inspiration from the principles of effective training data selection in supervised learning, particularly from the field of \textit{dataset condensation} \cite{DBLP_dataset_condensation}, we introduce Curriculum Latent Gradient (CLG), a novel approach for task-level in-context demonstration selection. CLG leverages the concept of \textit{latent task embeddings} and analyses their \textit{learning trajectory} to identify demonstrations that can effectively guide the model's learning process.

We begin by reviewing Latent Concept Learning (Sec.~\ref{sec:LCL}), highlighting its strengths and limitations as a foundation. We then articulate the core principles behind Curriculum Latent Gradient (Sec.~\ref{sec:LGA}), detailing how we capture and leverage the learning dynamics of latent task embeddings to guide demonstration selection. Finally, we describe the demonstration selection process based on these learning dynamics (Sec.~\ref{sec:DS}). An overview of our proposed methodology is depicted in Fig.~\ref{fig:main}.

\subsection{Preliminary: Latent Concept Learning}
\label{sec:LCL}
\paragraph{Latent Concept Learning} \citet{DBLP_latent_bayesian} aims to learn an optimal latent task concept $z_\theta$ as a proxy of in-context demonstrations. Specifically, prediction from a language model $P_M$ through in-context learning over demonstrations $D_n=\{(x_1,y_1),\dots,(x_n,y_n)\}$ can be expressed in Eq.~\ref{eq:icl},
\begin{equation}
\label{eq:icl}
    P(y \mid x; D_n) = P_M(y \mid x_1, y_1, \dots, x_n, y_n, x)
\end{equation}
where $D_n$ contains the $n$-shot demonstrations, and faces optimization challenges for its discrete nature. To address this, they replace the demonstration set with continuous and optimizable embeddings of the \textit{latent task concept} $z_\theta$, as expressed in Eq.~\ref{eq:latent_concept},
\begin{equation}
    \label{eq:latent_concept}
    P(y \mid x; z_\theta) = P_M(y \mid z_\theta, x)
\end{equation}
where $z_\theta$ can be optimized through maximum likelihood estimation on the entire training set $D_N$, as formulated in Eq.~\ref{eq:latent_concept_learning}.
\begin{equation}
    \label{eq:latent_concept_learning}
    \tilde{z_\theta} = \underset{z_\theta}{\arg\max} \sum_{(x, y) \in D_N} \log P(y \mid x;z_\theta)
\end{equation}

\paragraph{Latent-Bayesian} subsequently utilizes the optimized $\tilde{z_\theta}$ to select task-level demonstrations. Under a series of simplifying assumptions, it finally selects the top-$n$ demonstrations with the highest posterior probabilities of $\tilde{z_\theta}$ in language model $P_M$, as expressed in Eq.~\ref{eq:latent_bayesian}, where the $\mathrm{top}\text{-}n$ operator selects the $n$ samples with highest function values.
\begin{equation}
    \label{eq:latent_bayesian}
    D_n = \underset{(x,y) \in D_N}{\mathrm{top}\text{-}n} \ P_M(z_\theta \mid x, y)
\end{equation}

While Latent-Bayesian offers an interpretable approach under specific assumptions, its reliance on static posterior probabilities and independence assumptions limits its effectiveness, particularly for many-shot ICL. Crucially, it overlooks the \textit{dynamic interactions} between demonstrations – such as redundancy or synergistic effects – which become increasingly vital as the number of demonstrations grows. This motivates our departure from static, result-oriented approaches towards a method that explicitly considers the \textit{learning process} itself.

\subsection{Curriculum Latent Gradient}
\label{sec:LGA}
Our method borrows the idea of latent concept learning, i.e., learning task concept embeddings that play the role of context. Notably, we focus on the learning process instead of the learnt result. We quantify the learning dynamics under different demonstrations through the optimizing gradients on the task concept embeddings provided by each demonstration, throughout the training process.

Specifically, given a small language model $P_M$ and a target task, we construct $k$ prefix token embeddings $z_{\theta} \in R^{k \times h}$ to represent the latent task concept, where $h$ denotes the embedding size. These embeddings are then trained on the target task with the negative log-likelihood loss in Eq.~\ref{eq:nll}.
\begin{equation}
    \mathcal{L}(z_{\theta};x,y) = -\log{P_M(y|z_{\theta},x)}
    \label{eq:nll}
\end{equation}

We randomly initialize the latent concept tokens $z_{\theta} = z_{\theta_0}$, and employ Stochastic Gradient Descent \cite{DBLP_SGD} to optimize $z_{\theta}$ over the full training set $D_N$ for $E$ epochs. Crucially, we save the learnt latent concept at the end of each epoch, denoted as $z_{\theta_1}, z_{\theta_2}, \dots, z_{\theta_E}$. For each training example $(x,y) \in D_N$, we calculate its curriculum latent gradient, $G_{0:E}(x,y) \in R^{k \times h \times (E+1)}$, which is a concatenation of the gradients of the loss function with respect to $z_\theta$, evaluated both at the initial point $z_{\theta}=z_{\theta_0}$ and at the end of each epoch $z_{\theta}=z_{\theta_i}$ for $1 \leq i \leq E$. This is formulated in Eq.~\ref{eq:overall_latent_gradient}, where $\big[;\big]$ denotes the concatenation operation.
\begin{equation}
    \begin{aligned}
        G_{0:E}(x,y) = \big[&\nabla_{z_\theta} \mathcal{L}(z_\theta; x, y) \big|_{z_\theta = z_{\theta_0}}; \\
        &\nabla_{z_\theta} \mathcal{L}(z_\theta; x, y) \big|_{z_\theta = z_{\theta_1}}; \\
        &\dots; \\
        &\nabla_{z_\theta} \mathcal{L}(z_\theta; x, y) \big|_{z_\theta = z_{\theta_E}}\big]
    \end{aligned}
    \label{eq:overall_latent_gradient}
\end{equation}

\subsection{Demonstration Selection via Gradient Matching}
\label{sec:DS}
To select an effective $n$-shot demonstration set for ICL, we aim to identify an optimal $n$-shot subset $D_n^* \subseteq D_N$ that induces similar learning dynamics in the LM as the full training set $D_N$. Specifically, we propose to achieve this by minimizing the $L_2$ distance between the average curriculum latent gradient over $D_N$ and that over the $n$-shot subset $D_n$, as formalized in Eq.~\ref{eq:optimize_subset}, ensuring that the model's learning behaviour on the subset $D_n$, as summarized in $G_{0:E}(D_n)$, closely aligns with that on the complete set $D_N$, as summarized in $G_{0:E}(D_N)$.
\begin{equation}
    \begin{aligned}
        G_{0:E}(D_N) &= \mathbb{E}_{(x,y) \in D_N} \left[ G_{0:E}(x,y) \right] \\
        G_{0:E}(D_n) &= \mathbb{E}_{(x,y) \in D_n} \left[ G_{0:E}(x,y) \right] \\
        D_n^* &= \underset{D_n \subset D_N}{\arg\min} \, \left\| G_{0:E}(D_N) - G_{0:E}(D_n) \right\|_2
    \end{aligned}
    \label{eq:optimize_subset}
\end{equation}
\begin{algorithm}[t]
    \small
    \renewcommand{\algorithmicrequire}{\textbf{Input:}}
    \renewcommand{\algorithmicensure}{\textbf{Output:}}
    \caption{Gradient Matching through Greedy Search and Local Optimization}
    \label{alg:selection}
    \begin{algorithmic}[1]
        \REQUIRE the entire training set $D_N$, the curriculum latent gradient $G_{0:E}(x,y)$ for each example in the training set $(x,y) \in D_N$, the number of shots $n$, the maximum iteration steps $l$ for local optimization.
        \ENSURE the $n$-shot demonstrations $D_n^*$ that approximately minimize the gradient matching objective in Eq.~\ref{eq:optimize_subset}.
        
        \STATE Compute the average curriculum latent gradient on $D_N$: $G_{0:E}(D_N) = \mathbb{E}_{(x,y) \in D_N}\left[G_{0:E}(x,y)\right]$.
        
        \STATE Initialize $D_0 = \emptyset$.
        \FOR{$i = 1$ to $n$}
            \STATE Find the next example $(x_j, y_j) \in (D_N \setminus D_{i-1})$ that minimizes the distance between the average curriculum latent gradient on $D_N$ and that on $D_{i-1} \cup \{(x_j, y_j)\}$.
            \STATE Set $D_i = D_{i-1} \cup \{(x_j, y_j)\}$.
        \ENDFOR
        
        \STATE Set $d_\text{old} = \left\| G_{0:E}(D_N) - \mathbb{E}_{(x,y) \in D_n} G_{0:E}(x,y) \right\|_2$.
        \FOR{$i = 1$ to $l$}
            \STATE Find $(x_j, y_j) \in D_N \setminus D_n$ and $(x_k, y_k) \in D_n$ to minimize the distance between the average curriculum latent gradient on $D_N$ and that on $D_n \cup \{(x_j, y_j)\} \setminus \{(x_k, y_k)\}$.
            \STATE Set $D_{\text{new}} = D_n \cup \{(x_j, y_j)\} \setminus \{(x_k, y_k)\}$.
            \STATE Set $d_\text{new} = \left\| G_{0:E}(D_N) - \mathbb{E}_{(x,y) \in D_{\text{new}}} G_{0:E}(x,y) \right\|_2$.
            \IF{$d_\text{new} \geq d_\text{old}$}
                \STATE \textbf{break}
            \ELSE
                \STATE Set $D_n = D_{\text{new}}$ and $d_\text{old} = d_\text{new}$.
            \ENDIF
        \ENDFOR
        
        \STATE \textbf{return} $D_n$ as the optimized $n$-shot demonstrations $D_n^*$.
    \end{algorithmic}
\end{algorithm}

Since the optimization problem in Eq.~\ref{eq:optimize_subset} is indeed an NP-Complete Subset Sum Problem (SSP) \cite{DBLP_SSP}, we resort to heuristic approximation to find a near-optimal solution for $D_n^*$, involving two key phases: \textit{Greedy Search} and \textit{Local Optimization}. In the greedy phase, we incrementally construct the $n$-shot subset $D_n$ by iteratively adding the example that minimizes the target $L_2$ distance. Then the local optimization phase refines the $n$-shot subset through up to $l=32$ iterations to further reduce the target $L_2$ distance, by replacing a selected example with an unselected one. The detailed steps of the selection algorithm are presented in Algorithm~\ref{alg:selection}, which requires only a few minutes to execute in practice.

\section{Experiments}
We conduct experiments on 9 datasets from 5 distinct tasks, and show that demonstrations selected by our proposed method can effectively improve the many-shot ICL performances over random selection as well as some straightforward methods.

\subsection{Baselines}
In addition to random selection, the commonly-adopted method in existing work of many-shot ICL, we examine various straightforward methods for demonstration selection, including those designed for task-level selection and those extended from instance-level retrieval. We classify these methods in Table~\ref{tab:baselines} and introduce them in details below.
\begin{table}[t]
    \centering
    \small
    \setlength{\tabcolsep}{4pt}
    \begin{tabular}{lcc}
    \toprule
                   & \makecell{designed for\\task-level} & \makecell{extended from\\instance-level} \\
    \midrule
    learning-free  & \makecell{Random\\Best-of-N} & \makecell{BM25-Major\\BGE-KMeans} \\
    \midrule
    learning-based & \makecell{Latent-Bayesian\\CLG (ours)} & EPR-KMeans \\
    \bottomrule
    \end{tabular}
    \caption{Types of Methods}
    \label{tab:baselines}
\end{table}

\begin{table*}[t]
    \centering
    \small
    \setlength{\tabcolsep}{2.7pt}
    \begin{tabular}{llccccc}
        \toprule
        \textbf{Type} & \textbf{Dataset} & \textbf{Task} & \textbf{\#Train} & \textbf{\#Validation} & \textbf{Avg. Tokens} & \textbf{Metric} \\
        \midrule
        \multirow{4}{*}{Classification} 
        & SST-5 \cite{DBLP_sst5} & Sentiment Analysis & 8,544 & 1,101 & 26.55 & Acc \\
        & MNLI \cite{DBLP_mnli} & Natural Language Inference & 50,000 & 10,000 & 43.21 & Acc \\
        & CMSQA \cite{DBLP_cmsqa} & Commonsense Reasoning & 9,741 & 1,221 & 45.70 & Acc \\
        & HellaSwag \cite{DBLP_swag} & Commonsense Reasoning & 50,000 & 10,000 & 79.42 & Acc \\
        \midrule
        \multirow{5}{*}{Open-ended} 
        & GeoQuery \cite{DBLP_geoquery} & Code Generation & 600 & 280 & 22.66 & EM \\
        & NL2Bash \cite{DBLP_nl2bash} & Code Generation & 8,090 & 609 & 32.56 & BLEU \\
        & Break \cite{DBLP_break} & Semantic Parsing & 44,321 & 7,760 & 61.44 & LF-EM \\
        & MTOP \cite{DBLP_mtop} & Semantic Parsing & 15,667 & 2,235 & 34.12 & EM \\
        & SMCalFlow \cite{DBLP_smcalflow} & Semantic Parsing & 50,000 & 10,000 & 53.78 & EM \\
        \bottomrule
    \end{tabular}
    \caption{Datasets used in our experiments. We use at most 50,000 training instances and 10,000 validation instances. Following previous work \cite{DBLP_ceil}, we use Accuracy (Acc) on classification tasks, and use Exact Match (EM), Logical Form EM (LF-EM) \cite{DBLP_LFEM}, and character-level BLEU on open-ended tasks.}
    \label{tab:datasets}
\end{table*}

\paragraph{Random:} The most basic demonstration selection method, which is widely-adopted in existing work for many-shot ICL \cite{DBLP_manyshoticl,DBLP_manyshoticl_cls,DBLP_manyshoticl_verifier,DBLP_manyshoticl_multimodal}. We implement this across 5 random seeds and report the mean and standard deviation.

\paragraph{BM25-Major:} BM25 \cite{DBLP_bm25} is a popular term-based scoring method for instance-level retrieval \cite{DBLP_icl_bm25}. We extend it to task-level demonstration selection through Majority Voting, i.e., selecting $n$-shot demonstrations with the $n$ highest average scores to be retrieved by the other examples on the training set.

\paragraph{BGE-KMeans:} BGE-M3 \cite{DBLP_BGE_m3} provides off-the-shelf sentence embeddings with state-of-the-art performances on multiple retrieval tasks. We extend it to task-level selection through KMeans clustering, as implemented in the Scikit-Learn library, over its text embeddings.

\paragraph{EPR-KMeans:} EPR \cite{DBLP_icl_retrieval2} learns to retrieve demonstrations through contrastive learning, supervised by the ICL likelihoods on a relatively small LM. We train EPR retrievers and apply them to task-level selection through KMeans clustering over their learnt dense embeddings.

\paragraph{Best-of-N:} Best-of-N is a commonly used baseline for Reinforcement Learning (RL), and demonstrates competitive performance to RL-based task-level demonstration selection \cite{DBLP_icl_rl}. We implement this through randomly selecting N=5 demonstration sets, evaluating them on a relatively small LM over all training instances, and selecting the best-performing one.

\paragraph{Latent-Bayesian:} The task-level demonstration selection method based on Bayesian inference \cite{DBLP_latent_bayesian}, as we introduce in Sec~\ref{sec:LCL}.

\begin{table*}[ht]
\small
\centering
\setlength{\tabcolsep}{5pt}
\begin{tabular}{lcccccccccc}
    \hline
    \textbf{Method} & \textbf{SST-5} & \textbf{MNLI} & \textbf{CMSQA} & \textbf{HeSwag.} & \textbf{GeoQ.} & \textbf{NL2Bash} & \textbf{Break} & \textbf{MTOP} & \textbf{SMCal.} & \textbf{Average} \\
    \hline
    \textit{Llama3-70B} \\
    Random & 43.71 & 61.83 & \textbf{84.73} & \underline{76.11} & 73.36 & \underline{29.72} & 35.55 & 44.66 & 36.68 & 54.04 \\
    BM25-Major & 43.69 & 51.36 & 82.23 & 72.01 & 43.21 & 27.92 & 15.05 & 8.41 & 14.38 & 39.81 \\
    BGE-KMeans & 46.41 & 65.09 & 83.78 & 73.90 & \textbf{84.64} & 28.39 & 35.13 & 44.79 & \underline{38.73} & \underline{55.65} \\
    Best-of-N & 43.69 & \underline{70.92} & 84.19 & 74.57 & 80.71 & \textbf{29.76} & 35.86 & 43.40 & 37.66 & 55.64 \\
    EPR-KMeans & 43.05 & 49.78 & 83.62 & 71.32 & 78.21 & 22.92 & \textbf{38.39} & \textbf{50.43} & 38.10 & 52.87 \\
    Latent-Bayesian & \underline{46.59} & 69.84 & \underline{84.60} & 70.09 & 62.14 & 29.20 & 32.59 & 41.21 & 23.13 & 51.04 \\
    CLG (ours) & \textbf{48.32} & \textbf{76.37} & 84.52 & \textbf{80.77} & \textbf{84.64} & 29.59 & \underline{37.33} & \underline{47.74} & \textbf{40.35} & \textbf{58.85} \\
    \hline
    \textit{Qwen2.5-72B} \\
    Random & 36.00 & 58.20 & 87.47 & 86.58 & 57.29 & 33.28 & 36.67 & 45.32 & 38.23 & 53.23 \\
    BM25-Major & \underline{37.78} & 45.94 & 86.65 & 85.61 & 52.50 & 38.11 & 22.81 & 16.24 & 11.57 & 44.13 \\
    BGE-KMeans & 37.24 & 49.65 & \underline{87.71} & 86.67 & 61.79 & \underline{40.86} & 35.84 & 43.71 & \textbf{41.29} & 53.86 \\
    Best-of-N & 36.60 & \underline{60.57} & 87.55 & \underline{86.83} & 61.07 & 37.13 & 35.73 & 46.89 & 39.22 & \underline{54.62} \\
    EPR-KMeans & 36.15 & 58.41 & \textbf{87.96} & 86.22 & \textbf{62.86} & 26.53 & \textbf{39.12} & \textbf{47.53} & 39.65 & 53.83 \\
    Latent-Bayesian & 27.25 & 47.96 & 85.75 & 86.72 & 27.50 & 37.36 & 35.35 & 5.23 & 22.61 & 41.75 \\
    CLG (ours) & \textbf{38.33} & \textbf{77.29} & \underline{87.71} & \textbf{87.68} & \underline{62.50} & \textbf{45.19} & \underline{37.07} & \underline{47.07} & \underline{40.16} & \textbf{58.11} \\
    \hline
\end{tabular}
\caption{Main results (in \%) of Llama3-70B and Qwen2.5-72B under the 128-shot setting. The best scores on each model are in bold, and the second-best ones are underlined. Full results are illustrated in Appendix~\ref{sec:full_results}. BM25-Major and Latent-Bayesian generally perform worse than Random selection; we analyse the reasons in Appendix~\ref{sec:bm25_and_latent_bayesian}.}
\label{tab:main_results_concat_70B}
\end{table*}
\subsection{Datasets and Evaluation}
We list all the datasets in Table~\ref{tab:datasets}, paired with their tasks, sizes, average tokens, and evaluation metrics. We illustrate cases on each dataset in Appendix~\ref{sec:dataset_examples}. We select demonstrations from the training set and report their ICL performances on the validation set, since the test set is private for some datasets.

\subsection{Implementation Details}
For LM-based approaches, we utilize Llama3-8B to select demonstrations for Llama3 series LLMs \cite{DBLP_llama3}, and utilize Qwen2.5-3B to select demonstrations for Qwen2.5 series LLMs \cite{DBLP_Qwen25} and closed-source LLMs. We train EPR retrievers from BERT-base-uncased \cite{DBLP_bert} under the supervision of LMs, using a learning rate of $1e-5$ and a batch size of $8$, for at most $120$ epochs until regression. We conduct latent concept learning on LMs for Latent-Bayesian and CLG, with a learning rate of $1e-3$ and a batch size of $64$ for $E=10$ epochs.

For ICL performance evaluation, we utilize the vLLM framework \cite{DBLP_vLLM} to perform greedy search, involving 4-shot, 8-shot, 16-shot, 32-shot, 64-shot, and 128-shot settings.\footnote{Since the context window sizes of Llama3-8B and Llama3-70B are limited to $8192$, some of the 128-shot demonstrations may be truncated to fewer shots, ensuring at most $7000$ prompt tokens (the demonstrations and the test query).} On classification tasks, we only allow the model to generate tokens contained in the options.

\begin{figure*}[htp]
    \centering
    \includegraphics[width=\textwidth]{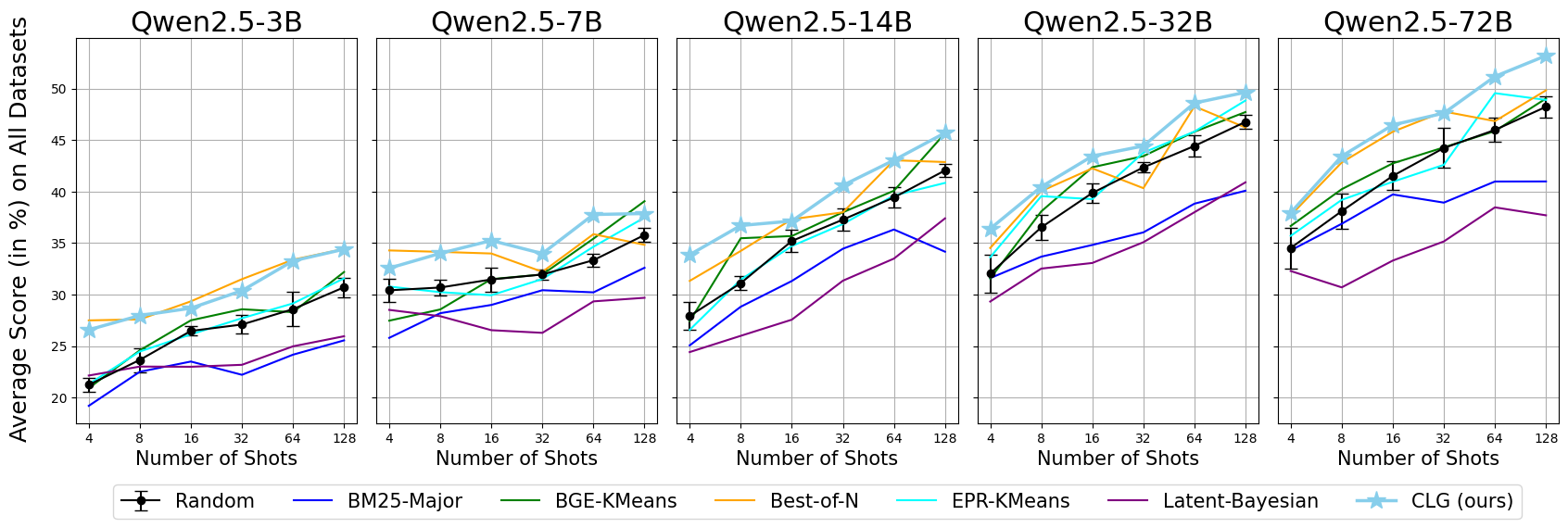}
    \caption{The scaling performances with respect to (w.r.t.) the number of shots and the model size. For random selection under 5 different random seeds, standard deviations are illustrated through error bars.}
    \label{fig:scaling}
\end{figure*}

\subsection{Main Results}
We illustrate the results of 128-shot ICL on Llama3-70B and Qwen2.5-72B in Table~\ref{tab:main_results_concat_70B}. It can be observed that, demonstrations selected by random are in fact generally suboptimal for task-level in-context learning. Among all methods, our proposed CLG performs the best on average, surpassing that of random selection by $4\%$ on both models.

\begin{table}[ht]
\small
\setlength{\tabcolsep}{4pt}
\centering
\begin{tabular}{lrrrrr}
    \hline
    \textbf{Method} & Qwen. & GLM4 & DouB. & GPT4o & DeepS. \\
    \hline
    Random     & 60.08 & 56.17 & 62.78 & 60.22 & 66.83 \\
    $\pm$std   & $\pm$0.22 & $\pm$2.24 & $\pm$0.26 & $\pm$0.45 & $\pm$0.52 \\
    Best-of-N  & 59.63 & 57.65 & 61.69 & 60.38 & 66.79 \\
    CLG (ours) & \textbf{61.87} & \textbf{58.50} & \textbf{64.09} & \textbf{61.48} & \textbf{68.08} \\
    \hline
\end{tabular}
\caption{ICL performances of 128-shot demonstrations, average over 9 datasets, on Qwen-turbo, GLM-4-flash, Doubao-pro, GPT-4o-mini, and DeepSeek-V3. Best scores are in bold. More details are in Appendix~\ref{sec:full_results_closed_source}.}
\label{tab:avg_results_closed_source}
\end{table}
We further evaluate the selected demonstrations by Random, Best-of-N, and CLG on 5 closed-source models: Qwen-turbo, GLM-4-flash, Doubao-pro, GPT-4o-mini, and DeepSeek-V3.\footnote{\url{https://www.alibabacloud.com/help/en/model-studio/getting-started/models}, \url{https://bigmodel.cn/dev/howuse/model}, \url{https://www.volcengine.com/product/doubao}, \url{https://openai.com/index/gpt-4o-mini-advancing-cost-efficient-intelligence}, and \url{https://api-docs.deepseek.com/news/news1226}.} We include at most 300 instances from each dataset and constrain the context length to within 7000 tokens due to budget limitations. The average performances across all datasets are illustrated in Table~\ref{tab:avg_results_closed_source}, where CLG consistently outperforms Random and Best-of-N on all closed-source models. This demonstrates the transferability of demonstrations selected by our CLG to different series of LLMs, as well as the effectiveness of CLG on closed-source LLMs.

\subsection{Overall Scaling Performance}
In Fig.~\ref{fig:scaling}, we examine the scaling trends of ICL performance w.r.t. the model size and the shot number. For simplicity, we present the average performance scores over 9 datasets. The results reveal a clear positive correlation between the overall ICL performance and the model size as well as the shot number, indicating that larger models and more shots generally lead to better performances.

Among the methods compared, CLG generally outperforms other methods across various conditions, achieving significant improvements (far more than the standard deviation) over random selection. In contrast, the most competitive Best-of-N baseline exhibits relatively lower scaling efficiency with the increased model size or the increased shot numbers. This further highlights the scalability of CLG with respect to the model size and the shot number.

\subsection{Scaling to a Thousand Shots}
To further compare CLG against Random selection in long-context settings with enough demonstrations, we select 1024-shot demonstrations through both methods, and evaluate their performances on MNLI, HellaSwag, Break, and SMCalFlow, the four datasets with the most examples.

\begin{table}[ht]
    \centering
    \small
    \begin{tabular}{lrrrr}
        \toprule
        \textbf{Method} & \textbf{MNLI} & \textbf{HeSwag.} & \textbf{Break} & \textbf{SMCal.} \\ \midrule
        \textit{128-shot} \\
        Random & 61.57 & 86.58 & 36.67 & 38.23 \\ 
        $\pm$std & $\pm$1.12 & $\pm$0.33 & $\pm$0.60 & $\pm$1.42 \\
        CLG (ours) & \textbf{77.29} & \textbf{87.68} & 37.07 & 40.16 \\
        $\Delta$ & +15.72 & +1.10 & +0.40 & +1.93 \\
        \midrule
        \textit{1024-shot} \\
        Random & 55.08 & 77.64 & 42.08 & 52.67 \\ 
        $\pm$std & $\pm$1.75 & $\pm$1.83 & $\pm$0.28 & $\pm$0.35 \\
        CLG (ours) & 58.94 & 79.12 & \textbf{42.21} & \textbf{52.82} \\ 
        $\Delta$ & +3.86 & +1.48 & +0.13 & +0.15 \\
        \bottomrule
    \end{tabular}
    \caption{Comparison of 128-shot and 1024-shot ICL on Qwen2.5-72B (in \%). Best scores are in bold.}
    \label{tab:1024-shot}
\end{table}
As illustrated in Table~\ref{tab:1024-shot}, CLG consistently achieves better results than the average performance of Random selection on all datasets. Such performance gains on 1024-shot are generally smaller than those on 128-shot. This could be explained by the possibility that 1024-shot demonstrations provide sufficient task knowledge needed for ICL, even when selected randomly.

Besides, we found that Break and SMCalFlow show significant improvements when scaled to 1024-shot, while MNLI and HellaSwag exhibit some degree of performance degradation. To explain this phenomenon, we analyse the scaling trends of ICL on each dataset with increasing shot numbers in Appendix \ref{sec:scaling_random_all_datasets}. We find that open-ended tasks, such as Break and SMCalFlow, are more likely to benefit from a larger number of shots.

\subsection{Ablation Study: Gradient Mismatching}
We conduct an ablation study to assess the effectiveness of the gradient matching term in Eq.~\ref{eq:optimize_subset}. Specifically, we reverse the optimization target: instead of minimizing the $L_2$ distance in Eq.~\ref{eq:optimize_subset}, we maximize it to select demonstrations that misalign with the full training set in learning dynamics.

\begin{figure}[htp]
  \includegraphics[width=\columnwidth]{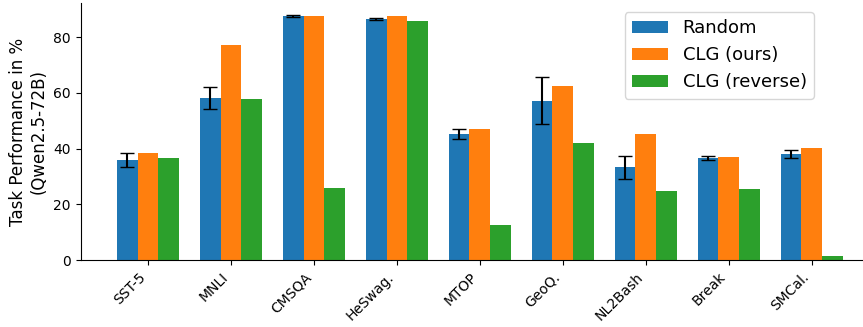}
  \caption{Ablation study on the impact of gradient mismatching on ICL performance, using 128-shot demonstrations with Qwen2.5-72B.}
  \label{fig:ablation}
\end{figure}

As illustrated in Fig.~\ref{fig:ablation}, this reversal results in significant degradation in the ICL performance, which can be much poorer than that of random selection. These results empirically validate that the quality of demonstrations is correlated with the degree of curriculum latent gradient matching.

\subsection{Improving Adaptability through Hybrid In-Context Learning}
In addition to our main experiments on task-level many-shot in-context learning, we further explore the complementary use of task-level and instance-level demonstrations. Specifically, we combine the 128-shot task-level demonstrations selected by Random and CLG with the 4-shot instance-level demonstrations retrieved by BGE-M3.

\begin{figure}[htp]
  \includegraphics[width=\columnwidth]{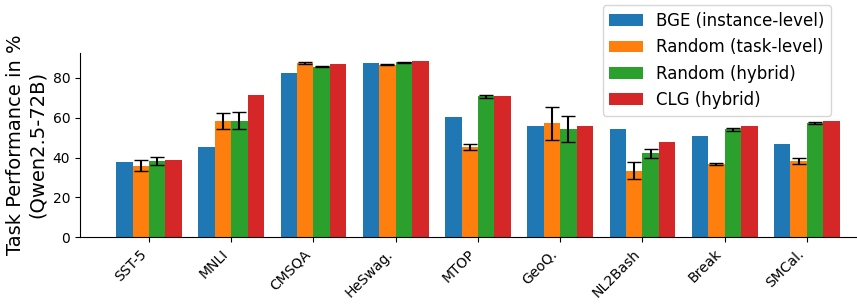}
\caption{ICL with hybrid demonstrations: task-level many-shot followed by instance-level few-shot.}
  \label{fig:hybrid}
\end{figure}

As illustrated in Fig.~\ref{fig:hybrid}, such hybrid strategy yields further improvements over using only task-level or instance-level demonstrations across most datasets. Meanwhile, CLG generally outperforms Random selection in such hybrid mode.

\subsection{Correlation between FT and ICL}
\begin{figure}[htp]
  \includegraphics[width=\columnwidth]{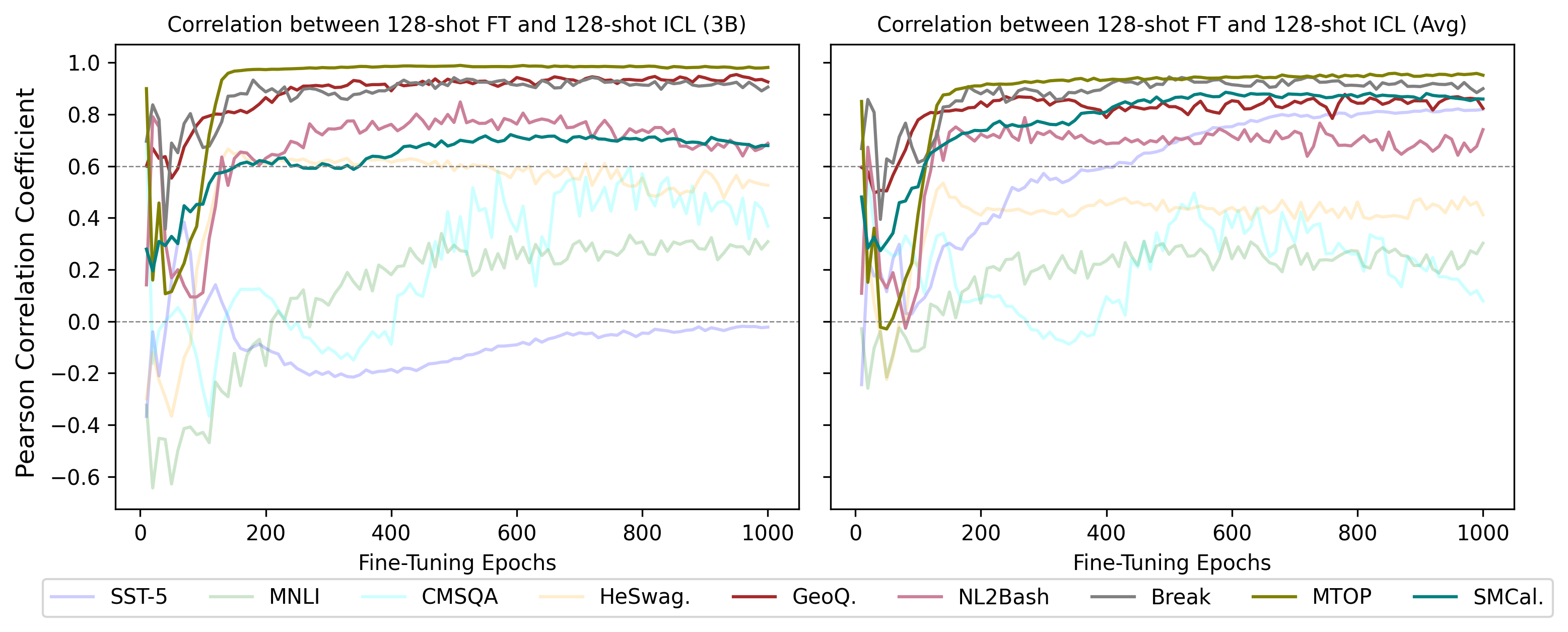}
\caption{The correlation between FT and ICL. The left involves ICL performances on Qwen2.5-3B, and the right involves that averaged over all Qwen2.5 models. Datasets for open-ended tasks are lined in solid colors.}
  \label{fig:ft_vs_icl}
\end{figure}

To validate our hypothesis that the data requirements for In-Context Learning (ICL) and Fine-Tuning (FT) are comparable, we fine-tune Qwen2.5-3B on the 128-shot demonstrations selected by each method. We employ prefix-tuning \cite{DBLP_prefix_tuning} to avoid overfitting, and perform training with a batch size of $128$ and a learning rate of $3e-4$ for $1000$ epochs. We quantify the FT performances through the negative evaluation losses on the validation set. We analyse and illustrate the correlation between FT and ICL performance in Fig.~\ref{fig:ft_vs_icl}, which shows a positive relationship between the two forms of machine learning, especially for open-ended tasks.

\subsection{Diversity and Coverage Analysis}
\label{sec:diversity_analysis}

Since prior work has illustrated that diversity and coverage are important for effective in-context demonstrations~\cite{DBLP_icl_diversity}, we examine whether our method inherently maintains diversity and coverage in selected demonstrations. Specifically, we analyze the label distributions of demonstration sets on classification datasets. We measure the KL divergence between the label distribution of selected demonstrations and that of the test set, with results shown in Table~\ref{tab:label_distribution}.

\begin{table}[ht]
    \centering
    \scriptsize
    \begin{tabular}{lrrrr}
        \toprule
        \textbf{Method} & \textbf{SST-5} & \textbf{MNLI} & \textbf{CMSQA} & \textbf{HeSwag.} \\
        \midrule
        Random & 0.020 & 0.016 & 0.016 & 0.012 \\
        $\pm$ std & $\pm$0.016 & $\pm$0.009 & $\pm$0.008 & $\pm$0.007 \\
        Best-of-N & 0.019 & 0.021 & 0.041 & 0.019 \\
        BGE-KMeans & 0.017 & 0.036 & 0.010 & 0.026 \\
        Latent-Bayesian & 4.050 & 0.115 & 0.134 & 4.296 \\
        CLG (ours) & \textbf{0.006} & \textbf{0.002} & \textbf{0.007} & \textbf{0.008} \\
        \bottomrule
    \end{tabular}
    \caption{KL divergence between demonstration set label distribution and test set label distribution (lower is better). CLG achieves the best alignment with test set distributions, indicating it inherently maintains diversity and coverage of labels on classification datasets.}
    \label{tab:label_distribution}
\end{table}

The results demonstrate that CLG naturally selects a demonstration set whose label distribution closely matches the test set distribution, significantly outperforming random selection. This suggests that gradient matching can inherently preserve data coverage in terms of label. In contrast, Latent-Bayesian shows particularly poor performance in maintaining label diversity, which may contribute to its suboptimal ICL performance.

\subsection{Efficiency Analysis}
\label{sec:efficiency}

To further verify the cost-effectiveness of our proposed CLG in practice, we analyse the computational cost of the complete demonstration selection pipeline. Table~\ref{tab:runtime} details the time consumption for selecting 128-shot demonstrations from MNLI and SMCalFlow using Qwen2.5-3B, measured on 8 A100 GPUs with data-parallel implementation via \texttt{deepspeed} and \texttt{accelerate} libraries.

\begin{table}[ht]
    \centering
    \small
    \begin{tabular}{lrrr}
        \toprule
        \textbf{CLG Stage} & \textbf{\# GPUs} & \textbf{MNLI} & \textbf{SMCal.} \\
        \midrule
        Prefix Tuning & 8 & 80 min & 87 min \\
        Gradient Computation & 8 & 145 min & 210 min \\
        Gradient Matching & 1 & 6 min & 10 min \\
        \bottomrule
    \end{tabular}
    \caption{Time consumption for 128-shot demonstration selection from datasets with 50k training examples.}
    \label{tab:runtime}
\end{table}

The results demonstrate that our approach maintains practical efficiency even at scale, with total computation time being approximately 3\textasciitilde4 times that of prefix-tuning, an already efficient fine-tuning method. This computational investment is well justified as our method enables significant ICL performance improvements on both open-source larger models and closed-source LLMs, for which fine-tuning may face challenge in computational resources and access to model weights. This validates the cost-effectiveness of our proposed CLG for many-shot ICL in practice.



\section{Conclusion}
Many-shot ICL has emerged as a promising way to utilize LLMs on downstream tasks, but the random selection applied in existing work may produce sub-optimal demonstrations and learning results.

In this work, we hypothesize that data requirements for in-context learning and fine-tuning are analogous, and propose Curriculum Latent Gradient (CLG) to select a demonstration set that aligns with the entire training set in learning dynamics on LMs. We validate our method across various datasets and LLMs, showing its significant improvements compared to random selection, e.g., improving by $4\%$ on open-source LLMs and by $2\%$ on closed-source LLMs.

This work unlocks more reliable and effective many-shot ICL, paving the way for its broader adoption and application in real-world scenarios.

\section*{Limitations}
This work focuses on improving many-shot in-context learning (ICL) by optimizing the demonstration set. However, ICL performance can also be sensitive to the order of demonstrations \cite{DBLP_icl_demo_order}, which our method does not address. We believe that investigating the impact of demonstration order on in-context learning could lead to further improvements in the learning performance, for example, by examining the learning dynamics at different stages of fine-tuning.

Besides, our method selects demonstrations according to analysis on the training set, which may suffer from biases in the training data, such as imbalances over gender, race, or culture. This could lead to discriminatory content in practice. Furthermore, improved many-shot ICL may be susceptible to malicious exploitation, resulting in misuse such as jailbreaking LLMs~\cite{google_manyshoticl_jailbreak}.

\section*{Acknowledgments}
This work was supported in part by the National Natural Science Foundation of China under Grant 62477001.

\bibliography{custom}
\clearpage
\appendix
\section{Dataset Examples}
\label{sec:dataset_examples}
We illustrate examples on all dataset in Table \ref{tab:case_demonstrations}. We illustrate a query prompt with random 4-shot demonstrations on each dataset, where the demonstration answers are highlighted in red.

\begin{table*}[h]
    \centering
    \fontsize{6pt}{8pt}\selectfont
    \begin{tabular}{lp{14cm}}
        \toprule
        Dataset & Query Prompt with 4-shot In-Context Demonstrations selected by Random \\
        \midrule
        SST-5 & more of the same from taiwanese auteur tsai ming-liang , which is good news to anyone who 's fallen under the sweet , melancholy spell of this unique director 's previous films . It is \textcolor{red}{good} \newline it 's a great performance and a reminder of dickens ' grandeur . It is \textcolor{red}{great} \newline a series of escapades demonstrating the adage that what is good for the goose is also good for the gander , some of which occasionally amuses but none of which amounts to much of a story . It is \textcolor{red}{bad} \newline one just waits grimly for the next shock without developing much attachment to the characters . It is \textcolor{red}{OK} \newline in his first stab at the form , jacquot takes a slightly anarchic approach that works only sporadically . It is  \\
        \midrule
        MNLI & Instead of ancient artefacts it shows the lifestyle and achievements of myriad Jewish communities around the globe through high-tec h audio-visual displays, hands-on exhibits, scale models (many of which are exquisite), and reconstructions. Can we say "The Jewish communities had many ancient artifacts on display."? \textcolor{red}{No} \newline Theirs is now the dominant right-wing critique of integrationist programs. Can we say "Right wing politics is praised by the integrationist."? \textcolor{red}{No} \newline Lorenzo the Magnificent and brother Giuliano lie in simple tombs beneath the sculptor's Madonna and Child, flanked by lesser artists' statues of the family patron saints Cosmas and Damian. Can we say "Lorenzo and Giuliano were related to one another."? \textcolor{red}{Yes} \newline The difference, if any, between the reacquisition price and the net carrying value of the extinguished debt should be recognized as a loss or gain in accounting for interest on Treasury debt. Can we say "The difference is not recognized."? \textcolor{red}{No} \newline The new rights are nice enough Can we say "Everyone really likes the newest benefits "?  \\
        \midrule
        CMSQA & Question: Bill sits down on a whoopee cushion, what sound does he make when he sits?    (A) fall asleep (B) flatulence (C) sigh of relief (D) medium (E) comfort        Answer: \textcolor{red}{B} \newline Question: What is likely heard by those going to a party?     (A) smoking pot (B) happiness (C) laughter (D) babies (E) meet new people       Answer: \textcolor{red}{C} \newline Question: A handsome prince is a stock character common to what?      (A) england (B) fairy tale (C) castle (D) palace (E) court      Answer: \textcolor{red}{B} \newline Question: What covers the largest percentage of the pacific northwest?       (A) united states (B) united states (C) washington (D) oregon (E) british columbia       Answer: \textcolor{red}{B} \newline Question: A revolving door is convenient for two direction travel, but it also serves as a security measure at a what?        (A) bank (B) library (C) department store (D) mall (E) new york Answer:  \\
        \midrule
        HeSwag. & Choose an ending: The player is holding a bat walking in the field. The player...     (A) in front stands across the screen and the ball dunks him to the ground. (B) makes a goal and the male scores. (C) runs down a purple net towards the ball. (D) begins to hit the ball.       Answer: \textcolor{red}{B} \newline Choose an ending: A woman sitting on the floor speaks to the camera. She...  (A) begins blowing cover on the phone. (B) pulls a harmonica and ties the end of the violin. (C) begins to play a saxophone in the bedroom. (D) holds onto a cat in front of her.        Answer: \textcolor{red}{D} \newline Choose an ending: Someone jumps out the living room window. He...     (A) dives for her pipe. (B) tuxedos her free eyes. (C) sees the girl chasing after him. (D) is as grim as he can.       Answer: \textcolor{red}{C} \newline Choose an ending: A group sits on horses as the stand and take a rest. The group...   (A) continues jumping around a gym as the crowd claps. (B) slowly slowly make their way to the other side of the pool on the level. (C) rides on the horses along a trail. (D) pass between horses at the left.  Answer: \textcolor{red}{C} \newline Choose an ending: Students lower their eyes nervously. She...        (A) pats her shoulder, then saunters toward someone. (B) turns with two students. (C) walks slowly towards someone. (D) wheels around as her dog thunders out.   Answer:  \\
        \midrule
        GeoQ. & how many rivers are there in m0 answer: \textcolor{red}{count(intersection(river,loc\_2(m0)))} \newline through which states does the m0 flow    answer: \textcolor{red}{intersection(state,traverse\_1(m0))} \newline through which states does the m0 run        answer: \textcolor{red}{intersection(state,traverse\_1(m0))} \newline what is the most populous city in m0       answer: \textcolor{red}{largest\_one(population\_1,intersection(city,loc\_2(m0)))} \newline name all the rivers in m0     answer:  \\
        \midrule
        NL2Bash & Get the path of running Apache        \textcolor{red}{ps -ef | grep apache} \newline Gets domain name from dig reverse lookup. \textcolor{red}{\$dig -x 8.8.8.8 | grep  PTR | grep -o google.*} \newline Find all *.xml files under current directory    \textcolor{red}{find -name *.xml} \newline Disables shell option 'nullglob'.     \textcolor{red}{shopt -u nullglob} \newline Add executable permission to "pretty-print"   \\
        \midrule
        Break & when did vincent von gogh die?  1\#) \textcolor{red}{return vincent von gogh 2\#) return when did \#1 die} \newline Does the Bronx have more non-Hispanic whites, or Hispanic whites?    1\#) \textcolor{red}{return the Bronx 2\#) return non-Hispanic whites in \#1 3\#) return Hispanic whites in \#1 4\#) return number of \#2 5\#) return number of \#3 6\#) return which is highest of \#4 , \#5} \newline What is the oldest log id and its corresponding problem id?        1\#) \textcolor{red}{return log ids 2\#) return \#1 that is the oldest 3\#) return corresponding problem id of \#2 4\#) return \#2 , \#3} \newline How many years apart were the British Saloon Car Championship season wins after 1970?  1\#) \textcolor{red}{return the British Saloon Car Championship 2\#) return season wins of \#1 3\#) return years of \#2 4\#) return \#3 that were after 1970 5\#) return the difference of \#4} \newline what flights are available tomorrow from denver to philadelphia  1\#) \\
        \midrule
        MTOP & Ask Rob how bad the traffic is   \textcolor{red}{[IN:SEND\_MESSAGE [SL:RECIPIENT Rob ] [SL:CONTENT\_EXACT how bad the traffic is ] ]} \newline Can I have the headlines please    \textcolor{red}{[IN:GET\_STORIES\_NEWS [SL:NEWS\_TYPE headlines ] ]} \newline what are the headlines in ohio news?        \textcolor{red}{[IN:GET\_STORIES\_NEWS [SL:NEWS\_TYPE headlines ] [SL:NEWS\_TOPIC ohio ] [SL:NEWS\_TYPE news ] ]} \newline What is the most recent news regarding local politics  \textcolor{red}{[IN:GET\_STORIES\_NEWS [SL:DATE\_TIME the most recent ] [SL:NEWS\_TYPE news ] [SL:NEWS\_CATEGORY local politics ] ]} \newline call Nicholas and Natasha   \\
        \midrule
        SMCal. & What about with Kaitlin Taylor?        \textcolor{red}{(Yield (Execute (NewClobber (refer (\^(Dynamic) ActionIntensionConstraint)) (\^(Recipient) ConstraintTypeIntension) (intension (RecipientWithNameLike (\^(Recipient) EmptyStructConstraint) (PersonName.apply "Kaitlin Taylor"))))))} \newline Sorry I meant the weekend? \textcolor{red}{(Yield (Execute (ReviseConstraint (refer (\^(Dynamic) roleConstraint (Path.apply "output"))) (\^(Event) ConstraintTypeIntension) (Event.start\_? (DateTime.date\_? (ThisWeekend))))))} \newline CHECK FOR ALBERT'S BIRTHDAY EVENT \textcolor{red}{(Yield (FindEventWrapperWithDefaults (Event.subject\_? (?~= "ALBERT'S BIRTHDAY"))))} \newline Add meetings same time everyday for this week.      \textcolor{red}{(FenceRecurring)} \newline Change the reservation for tonight to 6 people from 4.         \\
        \bottomrule
    \end{tabular}
    \caption{Dataset cases with random 4-shot in-context demonstrations.}
    \label{tab:case_demonstrations}
\end{table*}

\section{Case Study: Poor Performance of BM25-Major and Latent-Bayesian}
\label{sec:bm25_and_latent_bayesian}
Since BM25-Major and Latent-Bayesian perform poorly, even underperforming Random selection, we examine the demonstrations selected by these two methods. We found they tend to be trapped in specific patterns and lose diversity in selection. We illustrate their demonstration cases and discuss these phenomena in Tables~\ref{tab:case_demonstrations_latent_bayesian} and~\ref{tab:case_demonstrations_bm25_major} respectively.

\begin{table*}[ht]
    \centering
    \fontsize{6pt}{7pt}\selectfont
    \begin{tabular}{lp{14cm}}
        \toprule
        Dataset & Query Prompt with 4-shot In-Context Demonstrations selected by Latent-Bayesian \\
        \midrule
        SST-5 & renner ? It is \textcolor{red}{OK} \newline apart from its own considerable achievement , metropolis confirms tezuka 's status as both the primary visual influence on the animé tradition and its defining philosophical conscience . It is \textcolor{red}{OK} \newline ... salaciously simplistic . It is \textcolor{red}{OK} \newline ... would be a total loss if not for two supporting performances taking place at the movie 's edges . It is \textcolor{red}{OK} \newline in his first stab at the form , jacquot takes a slightly anarchic approach that works only sporadically . It is  \\
        \midrule
        HeSwag. & Choose an ending: Richard Parker frantically swipes at the small fish as someone shields his face. The fluttering silver creatures...       (A) spoon in around him. (B) inserting its wands into their palm. (C) pelt his exposed torso. (D) fly over him, catching and grabbing the sheet of flame.     Answer: \textcolor{red}{C} \newline Choose an ending: The sheer athleticism of his movement keeps him ahead of the choppers as he leads them on. Back at the window, someone...        (A) ducks down, gathers his cube and hunches at one. (B) groggily wakes with a crowbar. (C) gazes into a narrow gap between a tall arch. (D) leans her head against the rotting frame.        Answer: \textcolor{red}{D} \newline Choose an ending: The door closes after him, and unable to stop, the keys herd into it. Huge broken stone statues...       (A) lie on an maize field. (B) are about to emerge, which painted walls. (C) lie about the floor. (D) are drawn around the arena empty gleaming guns. Answer: \textcolor{red}{C} \newline Choose an ending: A strong wind flings leaves high into the air and the hedges advance. As the hedges close in behind them, they both...   (A) climb the rock - wakeboard on the side of the structure, to avoid it. (B) sip of the wand. (C) move, still throwing his legs to each other. (D) pose, once aftermath from all angles.     Answer: \textcolor{red}{D} \newline Choose an ending: Students lower their eyes nervously. She...      (A) pats her shoulder, then saunters toward someone. (B) turns with two students. (C) walks slowly towards someone. (D) wheels around as her dog thunders out.        Answer:  \\
        \bottomrule
    \end{tabular}
    \caption{Demonstration cases selected by Latent-Bayesian. On SST-5, Latent-Bayesian tends to exclusively select demonstrations with the neutral label \texttt{OK}. On HellaSwag, it predominantly selects demonstrations with answers \texttt{C} or \texttt{D}. Specifically, for 128-shot demonstrations on HellaSwag, Latent-Bayesian selected answers \texttt{A}, \texttt{B}, \texttt{C}, and \texttt{D} in proportions of $8$, $0$, $45$, and $75$ respectively. This is in contrast to the training set of $50,000$ examples where the four answers are distributed equally ($12,500 \pm 68$ each).}
    \label{tab:case_demonstrations_latent_bayesian}
\end{table*}

\begin{table*}[ht]
    \centering
    \fontsize{6pt}{7pt}\selectfont
    \begin{tabular}{lp{14cm}}
        \toprule
        Dataset & Query Prompt with 4-shot In-Context Demonstrations selected by BM25-Major \\
        \midrule
        Break & What shape is the object is a different shape and size than the other objects?  1\#) \textcolor{red}{return objects 2\#) return shapes of \#1 3\#) return sizes of \#1 4\#) return number of \#1 for each \#2 5\#) return number of \#1 for each \#3 6\#) return \#2 where \#4 is one 7\#) return \#3 where \#5 is one 8\#) return \#1 where \#2 is \#6 9\#) return \#1 where \#3 is \#7 10\#) return \#1 in both \#8 and \#9 11\#) return the shape of \#10} \newline Are all the spheres the same material?      1\#) \textcolor{red}{return spheres 2\#) return the materials of \#1 3\#) return number of \#1 for each \#2 4\#) return \#2 where \#3 is the highest 5\#) return \#1 where \#2 is \#4 6\#) return number of \#5 7\#) return number of \#1 8\#) return if \#6 and \#7 are equal} \newline What is the count and code of the job with the most employee?       1\#) \textcolor{red}{return employees 2\#) return jobs of \#1 3\#) return number of \#1 for each \#2 4\#) return the highest of \#3 5\#) return \#2 where \#3 is \#4 6\#) return the code of \#5 7\#) return \#4 , \#6} \newline What color is the object that is a different material than the others?      1\#) \textcolor{red}{return objects 2\#) return materials of \#1 3\#) return the number of \#1 for each \#2 4\#) return \#2 where \#3 is one 5\#) return \#1 where \#2 is \#4 6\#) return the color of \#5} \newline what flights are available tomorrow from denver to philadelphia         1\#) \\
        \midrule
        MTOP & remind me to call my father for fathers day      \textcolor{red}{[IN:CREATE\_REMINDER [SL:PERSON\_REMINDED me ] [SL:TODO [IN:GET\_TODO [SL:TODO [IN:CREATE\_CALL [SL:CONTACT [IN:GET\_CONTACT [SL:CONTACT\_RELATED my ] [SL:TYPE\_RELATION father ] ] ] ] ] [SL:TODO for fathers day ] ] ] ]} \newline Set a reminder to call my husband when I get home      \textcolor{red}{[IN:CREATE\_REMINDER [SL:TODO [IN:GET\_TODO [SL:TODO [IN:CREATE\_CALL [SL:CONTACT [IN:GET\_CONTACT [SL:CONTACT\_RELATED my ] [SL:TYPE\_RELATION husband ] ] ] ] ] [SL:LOCATION [IN:GET\_LOCATION ] ] ] ] ]} \newline Remind me to call my brother after work.\textcolor{red}{[IN:CREATE\_REMINDER [SL:PERSON\_REMINDED me ] [SL:TODO [IN:GET\_TODO [SL:TODO [IN:CREATE\_CALL [SL:CONTACT [IN:GET\_CONTACT [SL:CONTACT\_RELATED my ] [SL:TYPE\_RELATION brother ] ] ] ] ] [SL:TODO after work ] ] ] ]} \newline Remind me to call my father today.   \textcolor{red}{[IN:CREATE\_REMINDER [SL:PERSON\_REMINDED me ] [SL:TODO [IN:CREATE\_CALL [SL:CONTACT [IN:GET\_CONTACT [SL:CONTACT\_RELATED my ] [SL:TYPE\_RELATION father ] ] ] ] ] [SL:DATE\_TIME today ] ]} \newline call Nicholas and Natasha      \\
        \midrule
        SMCal. & What is on my calendar tomorrow? I want to have lunch with Sara, Barack and Monica.    \textcolor{red}{(do (Yield (FindEventWrapperWithDefaults (EventOnDate (Tomorrow) (\^(Event) EmptyStructConstraint)))) (Yield (CreateCommitEventWrapper (CreatePreflightEventWrapper (\& (\& (Event.subject\_? (?= "lunch")) (Event.start\_? (DateTime.date\_? (?= (Tomorrow))))) (Event.attendees\_? (\& (\& (AttendeeListHasRecipient (Execute (refer (extensionConstraint (RecipientWithNameLike (\^(Recipient) EmptyStructConstraint) (PersonName.apply "Sara")))))) (AttendeeListHasRecipient (Execute (refer (extensionConstraint (RecipientWithNameLike (\^(Recipient) EmptyStructConstraint) (PersonName.apply "Barack"))))))) (AttendeeListHasRecipient (Execute (refer (extensionConstraint (RecipientWithNameLike (\^(Recipient) EmptyStructConstraint) (PersonName.apply "Monica")))))))))))))} \newline I would like Tuesday. What time is my team sync with Abby, Elli and Kim? \textcolor{red}{(do (Yield (Execute (ChooseCreateEventFromConstraint (Event.start\_? (DateTime.date\_? (Date.dayOfWeek\_? (?= (Tuesday))))) (refer (\^(Dynamic) ActionIntensionConstraint))))) (Yield (Event.start (singleton (QueryEventResponse.results (FindEventWrapperWithDefaults (\& (Event.subject\_? (?~= "team sync")) (Event.attendees\_? (\& (\& (AttendeeListHasRecipientConstraint (RecipientWithNameLike (\^(Recipient) EmptyStructConstraint) (PersonName.apply "Abby"))) (AttendeeListHasRecipientConstraint (RecipientWithNameLike (\^(Recipient) EmptyStructConstraint) (PersonName.apply "Elli")))) (AttendeeListHasRecipientConstraint (RecipientWithNameLike (\^(Recipient) EmptyStructConstraint) (PersonName.apply "Kim"))))))))))))} \newline Change the lunch on Friday to being at 1pm. Also invite Sally, Alexa, and Sandra to the lunch.  \textcolor{red}{(Yield (UpdateCommitEventWrapper (UpdatePreflightEventWrapper (Event.id (singleton (QueryEventResponse.results (FindEventWrapperWithDefaults (\& (Event.subject\_? (?~= "lunch")) (Event.start\_? (DateTime.date\_? (Date.dayOfWeek\_? (?= (Friday)))))))))) (\& (Event.start\_? (DateTime.time\_? (?= (NumberPM 1L)))) (Event.attendees\_? (\& (\& (AttendeeListHasRecipient (Execute (refer (extensionConstraint (RecipientWithNameLike (\^(Recipient) EmptyStructConstraint) (PersonName.apply "Sally")))))) (AttendeeListHasRecipient (Execute (refer (extensionConstraint (RecipientWithNameLike (\^(Recipient) EmptyStructConstraint) (PersonName.apply "Alexa"))))))) (AttendeeListHasRecipient (Execute (refer (extensionConstraint (RecipientWithNameLike (\^(Recipient) EmptyStructConstraint) (PersonName.apply "Sandra"))))))))))))} \newline Put a lunch meeting with Abby, Kim, and Jesse at Taco Bell tomorrow        \textcolor{red}{(Yield (CreateCommitEventWrapper (CreatePreflightEventWrapper (\& (\& (\& (Event.subject\_? (?= "lunch meeting")) (Event.start\_? (DateTime.date\_? (?= (Tomorrow))))) (Event.location\_? (?= (LocationKeyphrase.apply "Taco Bell")))) (Event.attendees\_? (\& (\& (AttendeeListHasRecipient (Execute (refer (extensionConstraint (RecipientWithNameLike (\^(Recipient) EmptyStructConstraint) (PersonName.apply "Abby")))))) (AttendeeListHasRecipient (Execute (refer (extensionConstraint (RecipientWithNameLike (\^(Recipient) EmptyStructConstraint) (PersonName.apply "Kim"))))))) (AttendeeListHasRecipient (Execute (refer (extensionConstraint (RecipientWithNameLike (\^(Recipient) EmptyStructConstraint) (PersonName.apply "Jesse"))))))))))))} \newline Change the reservation for tonight to 6 people from 4.   \\
        \bottomrule
    \end{tabular}
    \caption{Demonstration cases selected by BM25-Major. BM25-Major tends to select significantly longer demonstrations. This can be attributed to the fact that longer examples tend to have higher BM25 scores.}
    \label{tab:case_demonstrations_bm25_major}
\end{table*}

\section{Full Results}
\label{sec:full_results}
We illustrate all results on Qwen2.5 series models and Llama3 series models for 128-shot scenarios in Table \ref{tab:main_results_concat_128}, and results averaged over 4-shot to 128-shot scenarios in Table \ref{tab:main_results_concat}.

\begin{table*}[ht]
\centering
\fontsize{8pt}{9pt}\selectfont
\begin{tabular}{lcccccccccc}
    \hline
    \textbf{Method} & \textbf{SST-5} & \textbf{MNLI} & \textbf{CMSQA} & \textbf{HeSwag.} & \textbf{GeoQ.} & \textbf{NL2Bash} & \textbf{Break} & \textbf{MTOP} & \textbf{SMCal.} & \textbf{Average} \\
    \hline
    \textit{Llama3-8B} \\
    Random & 37.75 & 54.32 & 72.99 & \underline{41.51} & 49.14 & 20.14 & 27.19 & 34.33 & 29.01 & 40.71 \\
    BM25-Major & 38.78 & 40.62 & 70.84 & 39.62 & 35.71 & \underline{28.66} & 2.16 & 5.37 & 6.56 & 29.81 \\
    BGE-KMeans & 37.15 & \underline{61.85} & 73.55 & 36.02 & \textbf{58.21} & 19.90 & 23.98 & 33.20 & 28.69 & 41.39 \\
    Best-of-N & \underline{39.33} & 51.92 & 73.38 & 40.58 & 53.93 & 26.78 & 25.88 & 35.48 & \textbf{30.55} & \underline{41.98} \\
    EPR-KMeans & 37.60 & 53.98 & \underline{73.87} & 39.99 & 49.64 & 17.65 & \textbf{30.09} & \underline{36.51} & \underline{29.63} & 41.00 \\
    Latent-Bayesian & 37.15 & 44.04 & 73.22 & 33.81 & 41.79 & 24.72 & 25.66 & 32.30 & 18.13 & 36.76 \\
    CLG (ours) & \textbf{39.96} & \textbf{61.98} & \textbf{73.96} & \textbf{46.01} & \textbf{58.21} & \textbf{33.83} & \underline{29.06} & \textbf{37.18} & 27.39 & \textbf{45.29} \\
    \hline
    \textit{Llama3-70B} \\
    Random & 43.71 & 61.83 & \textbf{84.73} & \underline{76.11} & 73.36 & \underline{29.72} & 35.55 & 44.66 & 36.68 & 54.04 \\
    BM25-Major & 43.69 & 51.36 & 82.23 & 72.01 & 43.21 & 27.92 & 15.05 & 8.41 & 14.38 & 39.81 \\
    BGE-KMeans & 46.41 & 65.09 & 83.78 & 73.90 & \textbf{84.64} & 28.39 & 35.13 & 44.79 & \underline{38.73} & \underline{55.65} \\
    Best-of-N & 43.69 & \underline{70.92} & 84.19 & 74.57 & 80.71 & \textbf{29.76} & 35.86 & 43.40 & 37.66 & 55.64 \\
    EPR-KMeans & 43.05 & 49.78 & 83.62 & 71.32 & 78.21 & 22.92 & \textbf{38.39} & \textbf{50.43} & 38.10 & 52.87 \\
    Latent-Bayesian & \underline{46.59} & 69.84 & \underline{84.60} & 70.09 & 62.14 & 29.20 & 32.59 & 41.21 & 23.13 & 51.04 \\
    CLG (ours) & \textbf{48.32} & \textbf{76.37} & 84.52 & \textbf{80.77} & \textbf{84.64} & 29.59 & \underline{37.33} & \underline{47.74} & \textbf{40.35} & \textbf{58.85} \\
    \hline
    \textit{Qwen2.5-3B} \\
    Random & 19.05 & 51.15 & 68.76 & 35.30 & 47.93 & 14.08 & 23.29 & 25.94 & 19.07 & 33.84 \\
    BM25-Major & \textbf{26.70} & 45.33 & 68.47 & 33.79 & 36.07 & \textbf{28.53} & 3.56 & 5.01 & 0.12 & 27.51 \\
    BGE-KMeans & \underline{26.43} & 40.70 & \textbf{73.79} & 32.72 & 51.43 & 19.74 & 19.83 & 26.26 & \textbf{22.20} & 34.79 \\
    Best-of-N & 21.25 & \textbf{56.02} & 71.50 & \underline{41.10} & \textbf{55.00} & \underline{21.36} & 23.45 & 26.31 & \underline{21.15} & \textbf{37.46} \\
    EPR-KMeans & 19.35 & \underline{54.53} & 70.93 & 33.61 & 49.29 & 11.79 & \underline{23.80} & \textbf{30.07} & 17.66 & 34.56 \\
    Latent-Bayesian & 21.07 & 44.66 & 69.62 & 25.68 & 28.93 & 11.26 & 20.91 & 23.80 & 11.56 & 28.61 \\
    CLG (ours) & 24.07 & 54.42 & \underline{72.56} & \textbf{44.77} & \underline{52.50} & 17.62 & \textbf{24.90} & \underline{26.49} & 19.40 & \underline{37.41} \\
    \hline
    \textit{Qwen2.5-7B} \\
    Random & \underline{26.48} & 31.88 & 84.88 & 71.38 & 35.43 & 16.93 & 28.22 & 32.72 & 27.99 & 39.55 \\
    BM25-Major & \textbf{29.34} & 31.88 & 83.78 & 71.50 & 41.43 & \textbf{32.27} & 8.65 & 7.79 & 11.98 & 35.40 \\
    BGE-KMeans & 26.25 & 31.88 & \underline{85.26} & \textbf{74.10} & \textbf{56.79} & 19.46 & 25.28 & \underline{33.20} & \textbf{30.73} & \textbf{42.55} \\
    Best-of-N & 24.80 & 31.88 & 84.77 & \underline{72.69} & 26.07 & 15.72 & 27.80 & 32.75 & \underline{28.71} & 38.35 \\
    EPR-KMeans & 24.70 & \textbf{31.94} & \underline{85.26} & 71.33 & \underline{51.79} & 15.44 & \textbf{31.17} & \textbf{35.30} & 25.43 & 41.37 \\
    Latent-Bayesian & 12.62 & 31.88 & 84.93 & 60.58 & 21.43 & 13.31 & 26.21 & 26.13 & 17.66 & 32.75 \\
    CLG (ours) & 24.89 & \textbf{31.94} & \textbf{85.34} & 71.76 & 50.00 & \underline{19.54} & \underline{28.27} & 32.98 & 28.03 & \underline{41.42} \\
    \hline
    \textit{Qwen2.5-14B} \\
    Random & 29.55 & \underline{56.76} & 82.29 & 55.53 & \underline{61.71} & 28.00 & 31.41 & 38.74 & 33.73 & 46.41 \\
    BM25-Major & 19.53 & 55.98 & 79.28 & 46.52 & 51.79 & \underline{41.82} & 14.85 & 10.02 & 12.82 & 36.96 \\
    BGE-KMeans & \textbf{34.79} & \textbf{57.31} & \underline{82.88} & 60.61 & \textbf{74.64} & 31.99 & 30.13 & 40.89 & 31.93 & \underline{49.46} \\
    Best-of-N & 28.97 & 55.17 & 82.56 & 52.92 & 59.64 & 36.30 & 30.44 & \underline{42.24} & 32.95 & 46.80 \\
    EPR-KMeans & 18.07 & 56.24 & 82.06 & 51.67 & 58.93 & 24.39 & \textbf{32.41} & \textbf{43.18} & \underline{33.95} & 44.54 \\
    Latent-Bayesian & 22.16 & 56.20 & 81.98 & \underline{63.08} & 47.50 & 24.77 & 30.39 & 28.95 & 16.71 & 41.30 \\
    CLG (ours) & \underline{32.79} & 56.30 & \textbf{83.05} & \textbf{69.35} & 60.36 & \textbf{42.43} & \underline{32.20} & 39.51 & \textbf{34.95} & \textbf{50.10} \\
    \hline
    \textit{Qwen2.5-32B} \\
    Random & 13.93 & 61.58 & 85.54 & 84.62 & 72.29 & 28.51 & 36.24 & 43.19 & 37.43 & 51.48 \\
    BM25-Major & 10.72 & 62.23 & 85.09 & 80.69 & 57.86 & \textbf{41.68} & 21.38 & 11.23 & 21.06 & 43.55 \\
    BGE-KMeans & 12.62 & 60.96 & 85.26 & 83.12 & \textbf{80.00} & 28.11 & 35.35 & 43.76 & \underline{38.51} & 51.97 \\
    Best-of-N & \textbf{17.44} & \underline{62.24} & 57.25 & \underline{85.31} & 73.57 & \underline{37.11} & 34.91 & \underline{46.94} & 35.36 & 50.01 \\
    EPR-KMeans & 14.62 & 61.41 & 85.67 & 83.60 & \underline{76.79} & 28.35 & \textbf{38.32} & \textbf{49.13} & \textbf{40.98} & \underline{53.21} \\
    Latent-Bayesian & 0.00 & 58.68 & \underline{85.91} & 81.66 & 58.57 & 31.57 & 34.88 & 33.24 & 22.21 & 45.19 \\
    CLG (ours) & \underline{16.17} & \textbf{73.33} & \textbf{86.40} & \textbf{86.53} & 73.57 & 29.97 & \underline{36.26} & 46.62 & 36.75 & \textbf{53.96} \\
    \hline
    \textit{Qwen2.5-72B} \\
    Random & 36.00 & 58.20 & 87.47 & 86.58 & 57.29 & 33.28 & 36.67 & 45.32 & 38.23 & 53.23 \\
    BM25-Major & \underline{37.78} & 45.94 & 86.65 & 85.61 & 52.50 & 38.11 & 22.81 & 16.24 & 11.57 & 44.13 \\
    BGE-KMeans & 37.24 & 49.65 & \underline{87.71} & 86.67 & 61.79 & \underline{40.86} & 35.84 & 43.71 & \textbf{41.29} & 53.86 \\
    Best-of-N & 36.60 & \underline{60.57} & 87.55 & \underline{86.83} & 61.07 & 37.13 & 35.73 & 46.89 & 39.22 & \underline{54.62} \\
    EPR-KMeans & 36.15 & 58.41 & \textbf{87.96} & 86.22 & \textbf{62.86} & 26.53 & \textbf{39.12} & \textbf{47.53} & 39.65 & 53.83 \\
    Latent-Bayesian & 27.25 & 47.96 & 85.75 & 86.72 & 27.50 & 37.36 & 35.35 & 5.23 & 22.61 & 41.75 \\
    CLG (ours) & \textbf{38.33} & \textbf{77.29} & \underline{87.71} & \textbf{87.68} & \underline{62.50} & \textbf{45.19} & \underline{37.07} & \underline{47.07} & \underline{40.16} & \textbf{58.11} \\
    \hline
\end{tabular}
\caption{Main results, merely 128-shot, on all models. The best scores on each model are in bold, and the second-best ones are underlined.}
\label{tab:main_results_concat_128}
\end{table*}

\begin{table*}[ht]
\centering
\fontsize{8pt}{9pt}\selectfont
\begin{tabular}{lcccccccccc}
    \hline
    \textbf{Method} & \textbf{SST-5} & \textbf{MNLI} & \textbf{CMSQA} & \textbf{HeSwag.} & \textbf{GeoQ.} & \textbf{NL2Bash} & \textbf{Break} & \textbf{MTOP} & \textbf{SMCal.} & \textbf{Average} \\
    \hline
    \textit{Llama3-8B} \\
    Random & 36.64 & 45.95 & 72.71 & 39.84 & 28.26 & 19.53 & 19.80 & 16.87 & 14.97 & 32.73 \\
    BM25-Major & 32.26 & 37.67 & 71.51 & 38.40 & 23.09 & 22.65 & 1.43 & 3.44 & 4.72 & 26.13 \\
    BGE-KMeans & 37.16 & 45.90 & 72.99 & 35.80 & 32.20 & 20.23 & 14.21 & 18.31 & 16.72 & 32.61 \\
    Best-of-N & \textbf{39.15} & \underline{48.50} & \textbf{73.63} & \underline{40.51} & \textbf{35.77} & \underline{27.22} & \underline{21.14} & \underline{20.42} & \underline{17.32} & \underline{35.96} \\
    EPR-KMeans & 35.12 & 43.26 & \underline{73.53} & 38.75 & 30.60 & 18.87 & 21.03 & 19.02 & 16.09 & 32.92 \\
    Latent-Bayesian & 36.69 & 37.88 & 72.66 & 34.15 & 18.15 & 16.15 & 18.96 & 14.21 & 9.79 & 28.74 \\
    CLG (ours) & \underline{38.72} & \textbf{51.86} & 73.52 & \textbf{44.27} & \underline{34.11} & \textbf{35.47} & \textbf{21.41} & \textbf{20.85} & \textbf{18.18} & \textbf{37.60} \\
    \hline
    \textit{Llama3-70B} \\
    Random & 42.39 & 50.82 & \underline{83.59} & 72.82 & 46.21 & 31.42 & 29.39 & 22.76 & 18.72 & 44.24 \\
    BM25-Major & 41.10 & 44.79 & 81.89 & 71.01 & 35.18 & 31.18 & 9.83 & 5.58 & 12.70 & 37.03 \\
    BGE-KMeans & 42.72 & \underline{53.00} & 82.90 & 60.84 & 50.06 & 32.34 & 27.11 & 22.67 & 19.89 & 43.50 \\
    Best-of-N & \underline{44.69} & 49.75 & 83.24 & \underline{75.63} & \textbf{51.79} & \underline{33.67} & 30.08 & 24.41 & \underline{21.77} & \underline{46.11} \\
    EPR-KMeans & 41.34 & 45.13 & 83.25 & 71.58 & 49.16 & 25.84 & \textbf{31.36} & \underline{26.56} & 21.40 & 43.96 \\
    Latent-Bayesian & 44.26 & 50.75 & 83.30 & 55.33 & 27.44 & 29.23 & 25.36 & 18.84 & 11.53 & 38.45 \\
    CLG (ours) & \textbf{46.17} & \textbf{57.28} & \textbf{83.69} & \textbf{79.48} & \underline{51.61} & \textbf{36.15} & \underline{30.47} & \textbf{27.18} & \textbf{24.30} & \textbf{48.48} \\
    \hline
    \textit{Qwen2.5-3B} \\
    Random & 17.81 & 49.79 & 70.77 & 36.96 & 25.67 & 15.98 & 14.48 & 13.18 & 9.80 & 28.27 \\
    BM25-Major & 21.15 & 50.18 & 68.40 & 36.65 & 20.83 & 18.32 & 1.01 & 2.26 & 0.16 & 24.33 \\
    BGE-KMeans & 21.69 & 46.76 & 72.37 & 34.03 & 28.87 & 17.23 & 9.93 & 13.23 & 12.18 & 28.48 \\
    Best-of-N & \textbf{23.28} & \textbf{60.32} & \underline{72.73} & \textbf{41.12} & \textbf{33.09} & \underline{19.18} & \underline{15.28} & 14.87 & \textbf{13.57} & \textbf{32.60} \\
    EPR-KMeans & 15.99 & 50.41 & 71.55 & 34.41 & \underline{30.89} & 12.99 & 14.14 & \textbf{16.41} & \underline{12.45} & 28.81 \\
    Latent-Bayesian & 20.85 & 51.93 & 67.98 & 32.97 & 13.21 & 10.81 & 13.37 & 9.50 & 10.20 & 25.65 \\
    CLG (ours) & \underline{22.09} & \underline{59.70} & \textbf{73.59} & \underline{39.39} & 27.14 & \textbf{27.20} & \textbf{16.61} & \underline{15.54} & 8.24 & \underline{32.17} \\
    \hline
    \textit{Qwen2.5-7B} \\
    Random & 23.84 & \underline{45.36} & 84.88 & \underline{75.65} & 14.69 & 17.38 & 21.92 & 16.09 & 14.81 & 34.96 \\
    BM25-Major & 24.51 & 41.30 & 83.58 & 73.83 & 17.86 & 20.58 & 5.44 & 3.70 & 8.76 & 31.06 \\
    BGE-KMeans & 24.95 & 32.15 & 84.98 & 75.48 & \underline{20.24} & 18.97 & 17.00 & 17.41 & 16.95 & 34.24 \\
    Best-of-N & \textbf{28.82} & 43.78 & \underline{85.04} & 75.38 & 15.53 & \underline{21.66} & \underline{22.18} & 18.16 & \underline{17.20} & \underline{36.42} \\
    EPR-KMeans & 19.94 & 44.04 & \textbf{85.16} & 75.38 & 18.81 & 15.30 & \textbf{23.80} & \textbf{19.63} & 15.31 & 35.26 \\
    Latent-Bayesian & 15.41 & 40.92 & 84.07 & 66.86 & 8.33 & 16.61 & 17.54 & 10.48 & 13.70 & 30.44 \\
    CLG (ours) & \underline{25.10} & \textbf{47.65} & 84.86 & \textbf{75.84} & \textbf{23.81} & \textbf{23.12} & 21.66 & \underline{18.50} & \textbf{17.53} & \textbf{37.56} \\
    \hline
    \textit{Qwen2.5-14B} \\
    Random & 25.67 & 52.56 & 76.58 & 64.15 & 34.23 & 30.25 & 24.96 & 19.53 & 17.27 & 38.35 \\
    BM25-Major & 16.09 & 48.36 & 78.31 & 61.59 & 36.61 & 37.25 & 8.92 & 6.42 & 8.16 & 33.52 \\
    BGE-KMeans & \underline{25.81} & \underline{52.96} & 77.98 & 64.23 & \underline{38.81} & 31.64 & 21.83 & 20.55 & 17.43 & 39.03 \\
    Best-of-N & 21.03 & 51.27 & 76.58 & 64.44 & \textbf{40.18} & \underline{40.16} & 25.57 & \underline{22.59} & \underline{18.81} & \underline{40.07} \\
    EPR-KMeans & 17.68 & 51.29 & 76.32 & 62.11 & 37.14 & 23.59 & \underline{25.67} & \textbf{23.32} & 18.19 & 37.26 \\
    Latent-Bayesian & 3.90 & 48.72 & \underline{78.33} & \underline{64.72} & 26.85 & 23.10 & 20.74 & 11.60 & 13.13 & 32.34 \\
    CLG (ours) & \textbf{29.32} & \textbf{53.83} & \textbf{79.06} & \textbf{70.19} & 35.24 & \textbf{41.38} & \textbf{25.79} & 21.93 & \textbf{20.44} & \textbf{41.91} \\
    \hline
    \textit{Qwen2.5-32B} \\
    Random & 12.85 & 62.43 & \underline{85.56} & 85.48 & 41.83 & 32.31 & \underline{28.97} & 22.12 & 19.01 & 43.40 \\
    BM25-Major & 11.52 & 58.64 & 85.56 & 83.73 & 34.46 & 35.52 & 12.34 & 5.60 & 15.12 & 38.06 \\
    BGE-KMeans & \textbf{20.78} & 57.03 & \textbf{85.63} & 84.82 & \textbf{49.40} & 31.00 & 26.03 & 21.89 & 19.29 & 43.99 \\
    Best-of-N & 17.94 & \textbf{66.03} & 71.27 & \underline{85.58} & 44.46 & \underline{40.06} & 28.53 & \underline{26.05} & 19.69 & 44.40 \\
    EPR-KMeans & 11.16 & 62.86 & 85.52 & 85.01 & \underline{48.63} & 31.08 & \textbf{29.82} & \textbf{26.16} & \underline{21.72} & \underline{44.66} \\
    Latent-Bayesian & 0.05 & 57.73 & 83.70 & 84.42 & 27.08 & 33.02 & 24.21 & 13.17 & 15.05 & 37.60 \\
    CLG (ours) & \underline{19.10} & \underline{63.38} & 85.46 & \textbf{86.67} & 44.46 & \textbf{41.55} & 28.01 & 25.46 & \textbf{22.66} & \textbf{46.31} \\
    \hline
    \textit{Qwen2.5-72B} \\
    Random & 34.13 & 57.59 & 87.12 & 87.73 & 32.49 & 35.95 & 29.40 & 23.37 & 19.67 & 45.27 \\
    BM25-Major & 28.76 & 49.16 & 85.72 & 86.88 & 36.25 & \underline{43.46} & 16.47 & 7.78 & 13.41 & 40.88 \\
    BGE-KMeans & \textbf{38.86} & 43.75 & \underline{87.24} & 87.36 & 40.18 & 40.63 & 27.29 & 25.22 & 21.57 & 45.79 \\
    Best-of-N & 37.54 & \underline{62.57} & 86.86 & 88.02 & 39.94 & 40.88 & \underline{30.37} & 25.14 & 20.98 & \underline{48.03} \\
    EPR-KMeans & 29.31 & 56.36 & \textbf{87.99} & 87.44 & \textbf{42.02} & 30.30 & \textbf{30.67} & \textbf{25.89} & \underline{21.85} & 45.76 \\
    Latent-Bayesian & 24.81 & 52.97 & 85.03 & \underline{88.14} & 15.42 & 34.75 & 25.17 & 3.31 & 15.17 & 38.31 \\
    CLG (ours) & \underline{38.30} & \textbf{64.05} & 87.03 & \textbf{88.32} & \underline{41.75} & \textbf{47.67} & 29.09 & \underline{25.63} & \textbf{23.79} & \textbf{49.51} \\
    \hline
\end{tabular}
\caption{Main results, averaged over 4-shot to 128-shot settings, on all models. The best scores on each model are in bold, and the second-best ones are underlined.}
\label{tab:main_results_concat}
\end{table*}

\section{Implementation Details and Full Results on Closed-Source LLMs}
\label{sec:full_results_closed_source}
We evaluate demonstrations selected by Random, Best-of-N, and CLG over 5 closed-source LLMs: Qwen-turbo, GLM-4-flash, Doubao-pro-32k, GPT-4o-mini, and DeepSeek-V3. These models are primarily designed for chat-based interactions with users rather than text completion. To adapt them for our evaluation, we begin by presenting them with a system message that outlines the task requirements:

\lstset{
    breaklines=true,
    basicstyle=\ttfamily\scriptsize,
    backgroundcolor=\color{white},
    frame=single,
    rulecolor=\color{gray},
    numbers=none,
}

\begin{lstlisting}
You are an In-Context Learner to learn from task demonstrations provided in context.
User will present you a prompt, composed of several demonstrations (query + response) and a test-query (only query).
You need to under stand the task concept and response format through the demonstrations, and return your response to the test query.

Example of a valid JSON response:
```json
{
    "response": "...",
}```
\end{lstlisting}

Next, we construct a user message containing the demonstrations (limited to 7000 tokens) and the test query, formatted as a JSON object. An example with 4-shot demonstrations is shown below:

\begin{lstlisting}
```json
{
    "demonstrations": [
        {
            "query": "as giddy and whimsical and relevant today as it was 270 years ago . It is ",
            "response": "great"
        },
        {
            "query": "after an uncertain start , murder hits and generally sustains a higher plateau with bullock 's memorable first interrogation of gosling . It is ",
            "response": "good"
        },
        {
            "query": "an unbelievably stupid film , though occasionally fun enough to make you forget its absurdity . It is ",
            "response": "bad"
        },
        {
            "query": "even though it is infused with the sensibility of a video director , it does n't make for completely empty entertainment It is ",
            "response": "OK"
        }
    ],
    "test-query": {
        "query": "in his first stab at the form , jacquot takes a slightly anarchic approach that works only sporadically . It is "
    }
}
```
\end{lstlisting}

After chatting with closed-source models by these messages, we extract and evaluate responses from their JSON-formatted outputs. The full evaluation results are illustrated in Table~\ref{tab:full_results_closed_source}.

\begin{table*}[ht]
\fontsize{8pt}{8pt}\selectfont
\centering
\begin{tabular}{lrrrrrrrrrr}
    \hline
    \textbf{Method} & \textbf{SST-5} & \textbf{MNLI} & \textbf{CMSQA} & \textbf{HeSwag.} & \textbf{GeoQ.} & \textbf{NL2Bash} & \textbf{Break} & \textbf{MTOP} & \textbf{SMCal.} & \textbf{Average} \\
    \hline
    \textit{Qwen-turbo-latest} \\
    Random & 56.87 & 80.93 & 86.47 & 79.07 & 73.43 & 62.61 & 44.20 & 28.73 & 28.40 & 60.08 \\
    $\pm$std & $\pm$2.36 & $\pm$0.80 & $\pm$1.21 & $\pm$0.44 & $\pm$3.97 & $\pm$1.15 & $\pm$2.88 & $\pm$2.22 & $\pm$1.99 & $\pm$0.22 \\
    Best-of-N & 52.67 & 80.33 & 85.33 & \textbf{80.00} & 72.86 & 61.79 & 42.00 & \textbf{34.00} & 27.67 & 59.63 \\
    CLG (ours) & \textbf{59.67} & \textbf{83.33} & \textbf{87.67} & 79.00 & \textbf{77.14} & \textbf{64.39} & \textbf{46.33} & 30.67 & \textbf{28.67} & \textbf{61.87} \\
    \hline
    \textit{GLM-4-flash} \\
    Random & 56.13 & 74.54 & \textbf{87.73} & 69.60 & 64.07 & 61.03 & 41.20 & 28.20 & 23.07 & 56.17 \\
    $\pm$std & $\pm$1.99 & $\pm$1.24 & $\pm$0.74 & $\pm$13.97 & $\pm$1.82 & $\pm$1.03 & $\pm$3.04 & $\pm$0.96 & $\pm$2.14 & $\pm$2.24 \\
    Best-of-N & 53.00 & \textbf{76.33} & 87.00 & 76.67 & \textbf{66.07} & 61.44 & 44.67 & \textbf{32.33} & 21.33 & 57.65 \\
    CLG (ours) & \textbf{56.33} & 72.67 & 87.33 & \textbf{77.67} & 63.93 & \textbf{62.24} & \textbf{51.67} & 30.67 & \textbf{24.00} & \textbf{58.50} \\
    \hline
    \textit{Doubao-pro-32k-0528} \\
    Random & 54.53 & 81.00 & 84.33 & \textbf{79.07} & 75.64 & \textbf{64.57} & 59.27 & 38.33 & 28.27 & 62.78 \\
    $\pm$std & $\pm$1.94 & $\pm$1.97 & $\pm$0.92 & $\pm$0.68 & $\pm$2.16 & $\pm$0.78 & $\pm$1.53 & $\pm$2.09 & $\pm$1.56 & $\pm$0.26 \\
    Best-of-N & 53.67 & 78.67 & \textbf{84.67} & 77.67 & 74.64 & 62.86 & 58.67 & 37.67 & 26.67 & 61.69 \\
    CLG (ours) & \textbf{55.00} & \textbf{81.33} & 83.67 & 77.67 & \textbf{79.64} & 62.82 & \textbf{60.33} & \textbf{46.33} & \textbf{30.00} & \textbf{64.09} \\
    \hline
    \textit{GPT-4o-mini} \\
    Random & 55.93 & 76.40 & 84.33 & 76.53 & 70.36 & 65.05 & 53.73 & 32.33 & \textbf{27.33} & 60.22 \\
    $\pm$std & $\pm$1.81 & $\pm$1.06 & $\pm$0.79 & $\pm$0.75 & $\pm$0.96 & $\pm$0.44 & $\pm$4.20 & $\pm$2.03 & $\pm$1.07 & $\pm$0.45 \\
    Best-of-N & 53.67 & 75.67 & 84.33 & \textbf{77.67} & \textbf{72.86} & 63.87 & 55.67 & 36.33 & 23.33 & 60.38 \\
    CLG (ours) & \textbf{56.00} & \textbf{78.67} & \textbf{86.67} & 76.67 & 68.93 & \textbf{65.72} & \textbf{57.33} & \textbf{36.67} & 26.67 & \textbf{61.48} \\
    \hline
    \textit{DeepSeek-V3} \\
    Random & \textbf{58.20} & 85.20 & 87.80 & 80.93 & 83.43 & \textbf{67.77} & 59.73 & 41.60 & 36.80 & 66.83 \\
    $\pm$std & $\pm$2.57 & $\pm$1.17 & $\pm$1.00 & $\pm$1.65 & $\pm$2.48 & $\pm$0.60 & $\pm$2.09 & $\pm$3.71 & $\pm$2.90 & $\pm$0.52 \\
    Best-of-N & 57.67 & \textbf{85.67} & 86.33 & 81.33 & 83.93 & 65.84 & 59.00 & \textbf{47.33} & 34.00 & 66.79 \\
    CLG (ours) & 56.33 & 85.33 & \textbf{88.00} & \textbf{83.33} & \textbf{87.14} & 67.28 & \textbf{62.00} & 45.33 & \textbf{38.00} & \textbf{68.08} \\
    \hline
\end{tabular}
\caption{Full results of closed-source models under the 128-shot setting. We only evaluate at most 300 instances on each dataset for limited budget. Best scores on each model are in bold.}
\label{tab:full_results_closed_source}
\end{table*}

\section{Scaling Trends of Random Selection on All Datasets}
\label{sec:scaling_random_all_datasets}
\begin{figure*}[!htbp]
    \centering
    \includegraphics[width=\textwidth]{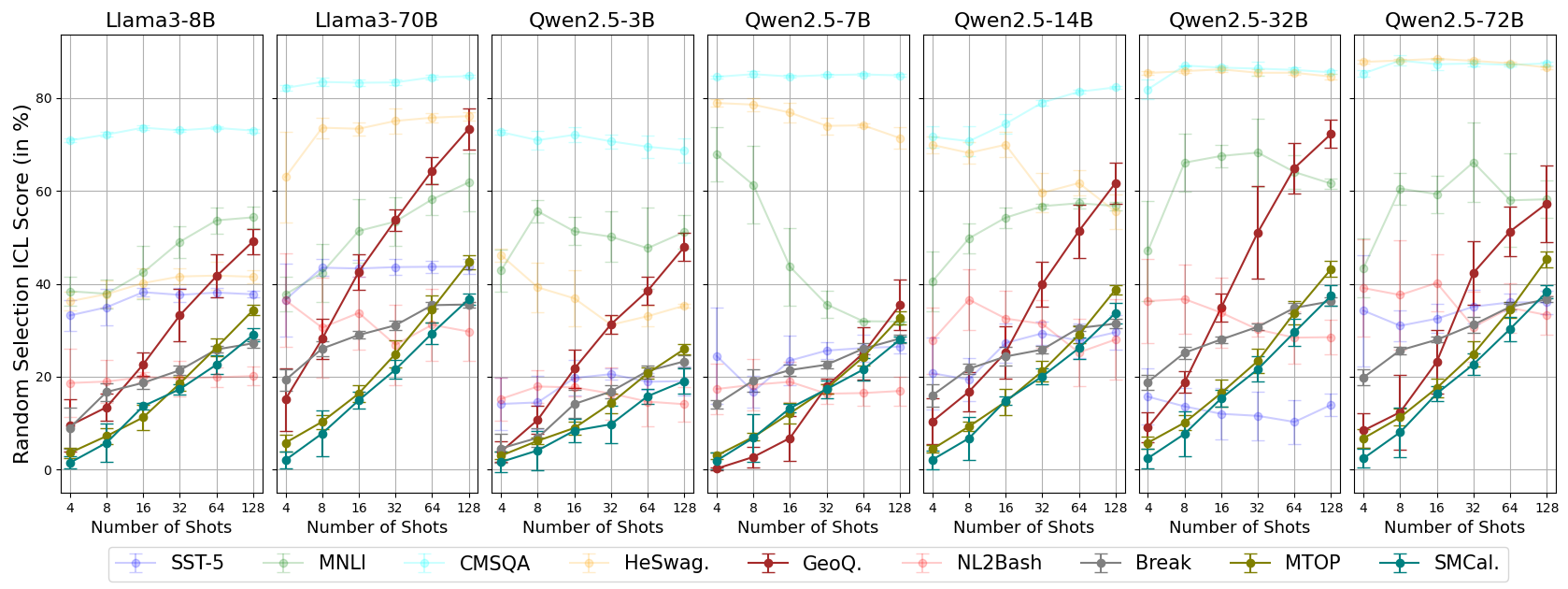}
    \caption{The scaling trends of random selection with respect to the number of shots on all models. We illustrate the standard deviations under 5 random seeds through error bars. The datasets that can steadily benefit from more shots are in solid colors, and the others are in semi-transparent colors.}
    \label{fig:scaling_random}
\end{figure*}

In Fig.~\ref{fig:scaling_random}, we present the ICL performances of all models across various demonstration sizes, ranging from 4-shot to 128-shot, using randomly selected demonstrations. The results indicate that datasets for open-ended tasks tend to consistently benefit from more in-context demonstrations. This trend is likely attributable to the fact that open-ended tasks have significantly longer answers (see Appendix~\ref{sec:dataset_examples}), which provide richer information to learn from context and more space for improvements in prediction.

\begin{figure*}[!htbp]
    \centering
    \includegraphics[width=0.7\textwidth]{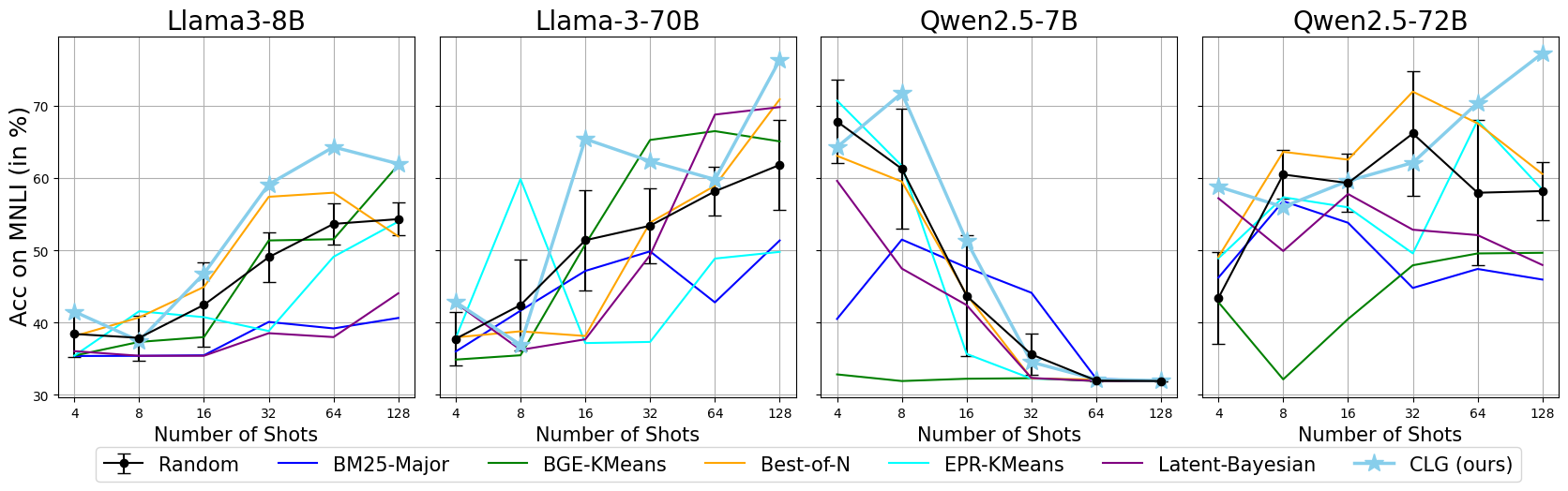}
    \caption{The performance of different models on MNLI.}
    \label{fig:qwen25_7b_mnli}
\end{figure*}

Qwen2.5-7B exhibits unusual performance on MNLI, significantly outperforming other models in the 4-shot setting but dropping to the lowest performance in the 128-shot setting. This anomaly is validated through comparisons across models and demonstration selection methods. As shown in Fig.~\ref{fig:qwen25_7b_mnli}, this behavior appears unique to Qwen2.5-7B and irrelevant to demonstration selection, suggesting Qwen2.5-7B may be a distinct case on MNLI.
\clearpage


\end{document}